\documentclass{article}

\usepackage{arxiv}

\usepackage[utf8]{inputenc} 
\usepackage[T1]{fontenc}    
\usepackage{hyperref}       
\usepackage{url}            
\usepackage{booktabs}       
\usepackage{amsfonts}       
\usepackage{nicefrac}       
\usepackage{microtype}      
\usepackage{lipsum}
\usepackage{graphicx}
\newcommand\ind{\hspace{\algorithmicindent}}
\usepackage{amssymb}
\usepackage{enumitem}
\usepackage{times}
\usepackage{algorithm}
\usepackage{algorithmic}
\usepackage{amsmath}
\usepackage{subcaption}
\usepackage{comment}
\usepackage{csquotes}
\usepackage{xcolor}
\usepackage{url}
\usepackage{tikz}
\usepackage{makecell}
\usepackage[toc,page]{appendix}

\graphicspath{ {./images/} }

\newcommand{\circlearrow}{%
  \mathrel{%
    \begin{tikzpicture}[baseline=-0.5ex]
      \node (circ) at (0,0) [circle,draw,inner sep=0pt,minimum size=0.4em] {};
      \draw[->] (circ.east) -- +(0.8em,0);
    \end{tikzpicture}%
  }%
}
\newcommand{\circlecircle}{%
  \mathrel{%
    \begin{tikzpicture}[baseline=-0.5ex]
      \node (leftcirc) at (0,0) [circle,draw,inner sep=0pt,minimum size=0.4em] {};
      \node (rightcirc) at (0.8em,0) [circle,draw,inner sep=0pt,minimum size=0.4em] {};
      \draw[-] (leftcirc.east) -- (rightcirc.west);
    \end{tikzpicture}%
  }%
}

\newcommand\mycite[1]{\textsuperscript{\cite{#1}}}

\title{CAnDOIT: Causal Discovery with Observational and Interventional Data from Time-Series}

\author{
 Luca Castri \\
  University of Lincoln\\
  \texttt{lcastri@lincoln.ac.uk} \\
   \And
 Sariah Mghames \\
  University of Lincoln\\
  \texttt{smghames@lincoln.ac.uk} \\
    \And
 Marc Hanheide \\
  University of Lincoln\\
  \texttt{mhanheide@lincoln.ac.uk} \\
  \And
 Nicola Bellotto \\
  University of Padua\\
  University of Lincoln\\
  \texttt{nbellotto@dei.unipd.it} \\
}

\begin{document}
\maketitle

\begin{abstract}
The study of cause-and-effect is of the utmost importance in many branches of science, but also for many practical applications of intelligent systems. In particular, identifying causal relationships in situations that include hidden factors is a major challenge for methods that rely solely on observational data for building causal models.
This paper proposes CAnDOIT, a causal discovery method to reconstruct causal models using both observational and interventional time-series data.
The use of interventional data in the causal analysis is crucial for real-world applications, such as robotics, where the scenario is highly complex and observational data alone are often insufficient to uncover the correct causal structure. Validation of the method is performed initially on randomly generated synthetic models and subsequently on a well-known benchmark for causal structure learning in a robotic manipulation environment. The experiments demonstrate that the approach can effectively handle data from interventions and exploit them to enhance the accuracy of the causal analysis. 

A Python implementation of CAnDOIT has also been developed and is publicly available on GitHub:~\url{https://github.com/lcastri/causalflow}.
\end{abstract}


\section{Introduction}
The increasing complexity of the environments in which future intelligent systems are expected to operate requires an understanding of the cause-effect relations between events, especially those affecting the performance of such systems.\mycite{pearl2009causality}
Unsurprisingly, \emph{Causal Inference} has become a crucial analysis across various research domains, including Earth science,\mycite{runge2023causal} healthcare,\mycite{saetia_constructing_2021} and robotics.\mycite{hellstrom2021relevance,scholkopf2021toward} To estimate the causal models that explain the available data, in recent decades many causal discovery algorithms for observational data in static and dynamic scenarios have been developed.\mycite{glymour_review_2019} That is, causal models are generated relying only on passive observations of the surrounding world, without taking into account the effect of hypothetical interventions. Observational data are often insufficient to retrieve the correct causal model in complex scenarios where it is impossible to account for all the variables responsible for the system's evolution. 
In such cases, hybrid approaches that incorporate first-principles into data-driven models can potentially improve the quality of the model.\mycite{zhu2019physics,raissi2019physics,gossler2017counterfactual,sivaram2022xai} Using first-principles models as constraints might reduce the degrees of freedom in causal discovery algorithms and improve the accuracy of their results.
However, this approach requires prior knowledge of the system's type or the fundamental physical laws governing its behavior, which can limit the algorithm's versatility. This potential solution has not been widely investigated in the causality community yet.
Conversely, an area that has been studied more extensively in the causality community involves using \emph{interventional data}, i.e., data from experiments, to eliminate spurious correlations and enhance the quality of the causal model.\mycite{pearl2009causality,rubin1974estimating} For this reason, recent works have led to the development of algorithms for static (i.e. time independent) data capable of constructing the causal model by leveraging both observations and interventions.\mycite{hauser2015jointly,brouillard2020differentiable,mooij2020joint,wang2022efficient,kocaoglu2019characterization,jaber2020causal,faria2022differentiable}
To our knowledge though, none of the previously mentioned works conduct causal discovery analysis for time-series data using observations and interventions. This represents a significant limitation for many real-world problems, such as sensor readings or dynamic process monitoring. Therefore, although it is obvious that using intervention information significantly enhances the quality of the causal model,\mycite{pearl2009causality,rubin1974estimating} our specific goal is to develop a causal discovery algorithm for time-series data capable of achieving this. Indeed, in this paper we extend and improve a recent state-of-the-art causal discovery for time-series data, LPCMCI,\mycite{gerhardus2020high} enabling it to retrieve the causal model from both observational and interventional data, which increases the accuracy of the causal analysis.
We named our solution CAnDOIT -- CAusal Discovery with Observational and Interventional data from Time-series.
%
The main contributions of this paper are therefore the following:
\begin{itemize}
    \item a new causal discovery algorithm for time-series capable of integrating interventional and observational data into the causal discovery process. To this end, we significantly enhance the accuracy of causal discovery, outperforming current state-of-the-art algorithm;
    \item a publicly available implementation of the algorithm, experimentally evaluated on various synthetic models from an ad-hoc random-model generator and a simulated robot manipulation scenario.
\end{itemize}

The paper is structured as follows: related work about causal from observations with and without interventions is presented in Section~\ref{sec:related_work}; Section~\ref{sec:approach} explains the implementation details of our method; Section~\ref{sec:randomdag} and Section~\ref{sec:causalworld_eval} present experimental results in randomly-generated and robot simulation models, respectively, to validate correctness and performance of our solution; finally, we conclude the paper in Section~\ref{sec:conclusion} discussing achievements and future work.

\section{Related Work}\label{sec:related_work}
\subsection{Observation-based Causal Discovery}
Structural causal models~(SCMs) and directed acyclic graphs~(DAGs) are at core of causal inference.\mycite{alma991011292629705181} They both represent system variables and their causal dependencies, the former with mathematical expressions, and the latter with nodes and oriented edges.
Various methods have been developed to derive causal relationships from observational data, categorized into three main classes:\mycite{glymour_review_2019} {\em (i) constraint-based methods}, such as Peter \& Clark~(PC) and Fast Causal Inference~(FCI),\mycite{spirtes2000causation} use conditional independence tests to recover the causal graph; {\em (ii) score-based methods}, like Greedy Equivalence Search~(GES)\mycite{10.1162/153244303321897717} and NOTEARS,\mycite{zheng2018dags} assigns scores to DAGs and explore the score space accordingly; {\em (iii) noise-based methods}, such as Linear Non-Gaussian Acyclic Models~(LiNGAM),\mycite{shimizu2006linear} identify causal structure within variables, assuming linearity, non-Gaussianity, and acyclicity of the model. 
However, most of these algorithms are applicable only to static data without temporal information, so they are not suitable for many interesting problems in Earth science,\mycite{runge2023causal} healthcare,\mycite{saetia_constructing_2021} or robotics.\mycite{hellstrom2021relevance,scholkopf2021toward} 

To address this limitation, various causal discovery algorithm have been developed to deal with time-series data.\mycite{assaad2022survey}
For example, within the area of Granger causality, the Temporal Causal Discovery Framework~(TCDF)\mycite{nauta2019causal} employs deep neural networks to learn complex nonlinear causal relationships among time-series. However, it faces challenges due to many hyperparameters and lacks a direct way to set the maximum time lags.
Among the noise-based methods, there are Time Series Models with Independent Noise~(TiMINo)\mycite{peters2013causal} and Vector Autoregressive LiNGAM (VARLiNGAM).\mycite{hyvarinen2010estimation} 
In the score-based methods class, the time-series version of NOTEARS, called DYNOTEARS,\mycite{pamfil2020dynotears} has been introduced.
%
In the constraint-based methods category, variations of the FCI and PC algorithms, namely time-series FCI~(tsFCI)\mycite{entner2010causal} and PC Momentary Conditional Independence~(PCMCI),\mycite{runge_causal_2018} were tailored to handle time-series data. PCMCI, in particular, has been applied to many research fields, including climate, healthcare, and robotics.\mycite{runge_detecting_2019,saetia_constructing_2021,castri2022causal} Recently, various extension of this algorithm have been introduced. 
PCMCI\textsuperscript{+},\mycite{runge2020discovering} for example, allows the discovery of simultaneous causal dependencies. Its further extension, Latent-PCMCI~(LPCMCI) allows simultaneous causal dependencies and latent confounders.\mycite{gerhardus2020high}
Filtered-PCMCI~(F-PCMCI),\mycite{castri2023enhancing} instead, includes an additional transfer entropy-based feature-selection module to remove unnecessary variables and perform causal discovery only on those responsible for the dynamic evolution of the observed scenario, resulting in faster model generation.
Joint-PCMCI\textsuperscript{+}~(J-PCMCI\textsuperscript{+})\mycite{pmlr-v216-gunther23a} enables causal structure learning from multiple observational datasets. 
Finally, PCMCI$_\Omega$,\mycite{gao2023causal} assumes a finite periodical repetition of causal mechanisms, which is suitable for semi-stationary time-series data.

Despite dealing with time-series, the above algorithms work only with observational data and cannot handle the theoretical concept of intervention, specifically the so-called \emph{do-operator}.\mycite{pearl2009causality} This operator overrides the structural equations of the intervention variables, disrupting the original dependencies in the observational setting. The concept of intervention is a fundamental aspect of causality. Studying how a variable reacts in response to a forced modification of another variable provides a deeper understanding of the system compared to what simple observations can offer.

\subsection{Observation and Intervention-based Causal Discovery}
Incorporating interventional data into the causal discovery process differs from its use in other causal problems, like measuring the Average Treatment Effect~(ATE).\mycite{rubin1974estimating} The ATE is a metric employed to evaluate the causal influence of a specific intervention on an outcome, given the presence of a causal model. Conversely, interventional data can be used to enhance the accuracy of the discovery process to find that model.
Formally, performing causal discovery on observational data means that the true DAG is only identifiable up to a Markov equivalence class~(\emph{MEC}).\mycite{pearl2009causality} However, by incorporating interventional data, identifiability can be improved up to an interventional class~($\mathcal{I}$\emph{-MEC}),\mycite{yang2018characterizing} which is a subset of \emph{MEC}. The same consideration applies when dealing with Maximal Ancestral Graphs~(MAGs), where the equivalence class is represented by a Partial Ancestral Graph~(PAG), capturing the conditional independence relations among observed variables while accounting for latent variables.
For this reason, various algorithms have been developed for causal discovery using both observational and interventional datasets.

The Interventional Greedy Sparsest Permutation~(IGSP)\mycite{yang2018characterizing} algorithm has been proposed to learn causal DAGs when both observational and interventional data are available. Its extension, Unknown Target IGSP~(UT-IGSP),\mycite{squires2020permutation} enables DAGs learning also when the intervention targets are partially or completely unknown.
An extension of the FCI causal discovery method has been proposed to enable causal model learning by combining observations and experiments (i.e. soft interventions).\mycite{kocaoglu2019characterization} This extension has been further improved by $\Psi$-FCI to handle unknown intervention targets.\mycite{jaber2020causal}
Recent neural approaches, such as  Differentiable Causal Discovery with Interventions~(DCDI)\mycite{brouillard2020differentiable} and Efficient Neural Causal Discovery~(ENCO),\mycite{lippe2021efficient} can handle various types of interventional data. An extension of DCDI has also been developed to use latent interventions.\mycite{faria2022differentiable}
%
%
The Multiple Interventional Datasets for Efficient Global causal Structure learning~(EMIDGS)\mycite{wang2022efficient} algorithm learns a causal skeleton using a variant of the PC algorithm adapted to handle multiple interventional datasets simultaneously. It then orients the skeleton by identifying an $\mathcal{I}$\emph{-MEC} and applies a score function to conduct a greedy search for each causal DAG from each dataset within the remaining search space. Finally, the graphs are merged to obtain the final causal structure.
Recently, a novel framework, called {\em Joint Causal Inference}~(JCI),\mycite{mooij2020joint} has been proposed to facilitate causal discovery considering different contexts of the same scenario. The authors elaborate on how this approach can be employed to model interventions as context changes. By exploiting this framework, existing observational causal inference algorithms can be enhanced to learn causal structures from interventional data.

All these works, however, conduct causal analysis from observations and interventions on static datasets. As already mentioned, this limitation poses significant challenges for many applications where temporal information cannot be ignored and is crucial for correctly modeling the system. 
Our approach targets these types of applications and differs from previous solutions as it performs causal discovery on time-series data by incorporating both observations and interventions.

\section{Causal Discovery Based on Observational and Interventional Data}\label{sec:approach}
The solutions proposed in this paper, called CAnDOIT, extends and improves the state-of-the-art causal discovery algorithm for time-series data, LPCMCI,\mycite{gerhardus2020high} taking inspiration from the way JCI\mycite{mooij2020joint} handles interventions with known target. The result is a new algorithm that enables precise causal analysis, using both observational and interventional data, with significantly improved accuracy. Specifically, CAnDOIT extends LPCMCI by performing causal discovery with hard interventions on known targets.

\subsection{LPCMCI}
The original LPCMCI\mycite{gerhardus2020high} algorithm is an enhancement of PCMCI\textsuperscript{+},\mycite{runge2020discovering} extending it by removing the causal sufficiency assumption as an initial condition. LPCMCI retains the ability to detect both time-lagged and contemporaneous cause-effect relationships between variables from its predecessors, PCMCI and PCMCI\textsuperscript{+}. Additionally, it relaxes the causal sufficiency assumption, allowing for the inclusion of latent confounders, i.e., unobserved common causes.

In more detail, LPCMCI is a constraint-based causal discovery algorithm that begins with a fully connected graph~$\mathcal{G}$. The algorithm then proceeds through a preliminary phase, which removes many (though not necessarily all) false links and repeatedly applies a set of orientation rules to define the heads and tails of the edges. These rules help identify a subset of the \mbox{(non-)ancestorships} in~$\mathcal{G}$. The identified non-ancestorships and ancestorships further constrain the conditioning sets for subsequent conditional independence (CI) tests. The goal of this iterative process is to accurately determine a subset of the parentships in $\mathcal{G}$. This subset is then used in the final phase of the algorithm, which involves further removal of links, application of orientation rules, and use of identified \mbox{(non-)ancestorships}. Ultimately, LPCMCI outputs a time-series PAG.

In real-world applications, the challenge of latent confounders is almost always present, as it is generally impossible to account for all variables influencing the system's evolution. Consequently, given our interest for real-world applications in dynamic scenarios, we based our CAnDOIT algorithm on LPCMCI to leverage its capability to handle hidden confounders effectively.

\subsection{Interventions through Context Variables}
Combining observational and interventional data in the causal discovery process requires the causal structure related to the intervention variable to adapt to both the observational and interventional cases. Specifically, we need to consider the parents of the intervention variable when dealing with observational data, while we must break all the links affecting the intervention variable when performing the intervention. To address this challenge, we use context nodes to model interventions, taking inspiration from the JCI framework. The latter distinguishes between {\em system variables}, which describe the actual system, and {\em context variables}, which refer to the observation context. While the system variables are treated as endogenous, the context variables are typically (but not necessarily) exogenous.
System and context variables form a new {\em meta-system} $\mathcal{M}$, which is used for the causal analysis.
Accordingly, our meta-system is defined as follows:

\begin{equation}\label{eq:metasys}
    \mathcal{M} : 
    \begin{cases}
        X_i(t) = \tilde{f}(P\!a(X_i), C\!X_{k}) & i \in \mathcal{I}, k \in \mathcal{K}\\
        C\!X_k = f_k & k \in \mathcal{K}
    \end{cases}
\end{equation}
where $\mathcal{I}$ represents the set of system variables defined as $X = (X_i)_{i\in\mathcal{I}}$, while $\mathcal{K}$ represents the set of context variables defined as $C = (C\!X_k)_{k\in\mathcal{K}}$. $P\!a(X_i)$ is the parent set of the system variable $X_i$, instead $C\!X_{k}$ is the context variable $k$. Moreover, the function $\tilde{f}$ models the system variables and can be decomposed as follows:
\begin{equation}
    \tilde{f}(P\!a(X_i), C\!X_{k}) : 
    \begin{cases}
        f(P\!a(X_i)) & C\!X_{k} = 0\\
        C\!X_{k} & C\!X_{k} = \xi_k
    \end{cases}
\end{equation}
where $f$ represents the function that models the evolution of the system variable $X_i$ in the observational case, which depends solely on its parent set $Pa(X_i)$. Referring again to \textbf{Equation~\ref{eq:metasys}}, the function $f_k$ models the context variables and it is defined as follows:
\begin{equation}
f_k : 
\begin{cases}
\xi_k & \text{interventional mode for $k$}\\
0 & \text{observational mode for $k$}
\end{cases}
\end{equation}
The case where $C\!X_{k}=0$ corresponds to no intervention, i.e the observational baseline, while $C\!X_{k}=\xi_k$ models the intervention case, where $\xi_k$ is the actual intervention value assigned to the variable $X_i$ through the context variable $C\!X_{k}$. This approach is consistent with the concept of \enquote{force variables} adopted in the literature.\mycite{mooij2020joint,pearl1993bayesian}
In our case, to model hard interventions with known targets as context changes in time-series data, we assume the three assumptions made by the JCI framework, specifically we used the \emph{JCI123} version of the framework.\mycite{mooij2020joint} 
The three assumptions are the following:
\begin{enumerate}[label={\textbf{JCI\arabic*}}, leftmargin=1.5cm]
    \item \emph{Exogeneity}: No system variable causes any context variable, i.e.,
    \[\forall k \in \mathcal{K}, \forall i \in \mathcal{I}: i \rightarrow k \notin \mathcal{G}(\mathcal{M})\]
    \item \emph{Complete randomised context}: No context variable is confounded with a system variable, i.e.,
    \[\forall k \in \mathcal{K}, \forall i \in \mathcal{I}: i \leftrightarrow k \notin \mathcal{G}(\mathcal{M})\]
    \item \emph{Generic context model}:  The context graph $\mathcal{G}(\mathcal{M})_{\mathcal{K}}$ is of the following special form:
    \[\forall k \neq k' \in \mathcal{K}: k \leftrightarrow k' \in \mathcal{G}(\mathcal{M}) \land k \rightarrow k' \notin \mathcal{G}(\mathcal{M})\]
\end{enumerate}
where $\mathcal{G}(\mathcal{M})$ represents the graph of the meta-system $\mathcal{G}$, and $\mathcal{G}(\mathcal{M})_{\mathcal{K}}$ is the sub-graph of $\mathcal{G}(\mathcal{M})$ specifically related to the context variables.
The \emph{exogeneity} assumption (JCI1) is commonly applied when the context encodes interventions on the system that have been predetermined and executed before any measurements are taken. This is the approach we use to model and conduct interventions. Assumption 2 posits that the choice of intervention is independent of any other factors that might influence the system, and that the observed context variables fully describe the context.
JCI3 is introduced to model a scenario where the context distribution contains no (conditional) independencies. This assumption is made when the goal is not to model causal relationships among the context variables themselves, but rather to use these variables to help model causal relationships among the system variables. Note that JCI3 can only be assumed if both JCI1 and JCI2 hold, which is true in our case.
Additionally, since we are dealing with interventions with known targets, we have added a fourth assumption: each context node can only influence a single system variable, specifically the one undergoing the intervention. This is because the intervention variable is known and provided to the algorithm.

\begin{figure}[t]
    \centering    
    \begin{subfigure}{.465\textwidth}
        \includegraphics[trim={0.2cm 5cm 7.5cm 0cm}, clip, width=\textwidth]{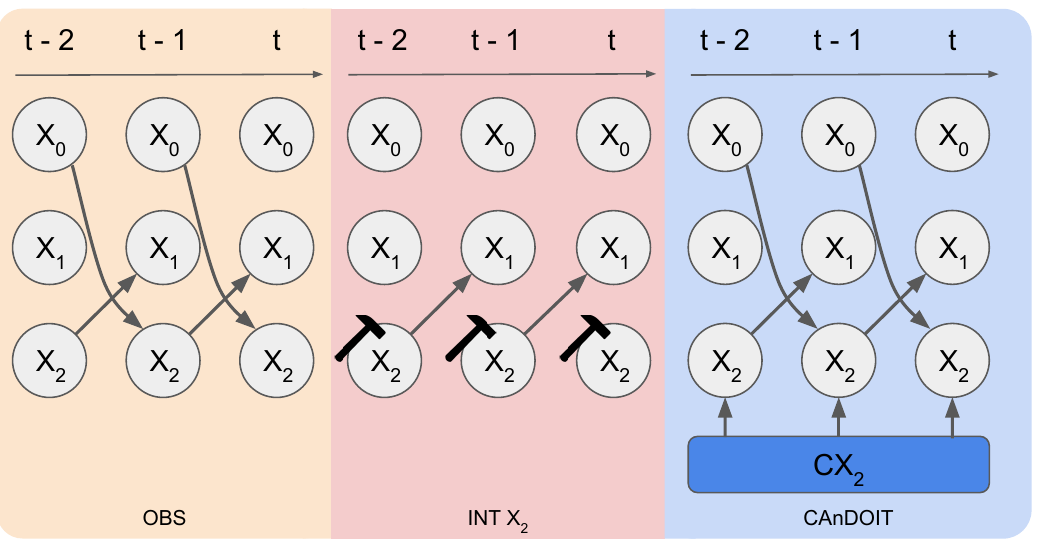}\caption{}\label{fig:approach_example_single}
    \end{subfigure}
    \begin{subfigure}{.53\textwidth}
        \includegraphics[trim={0.2cm 3.5cm 1.8cm 0cm}, clip, width=\textwidth]{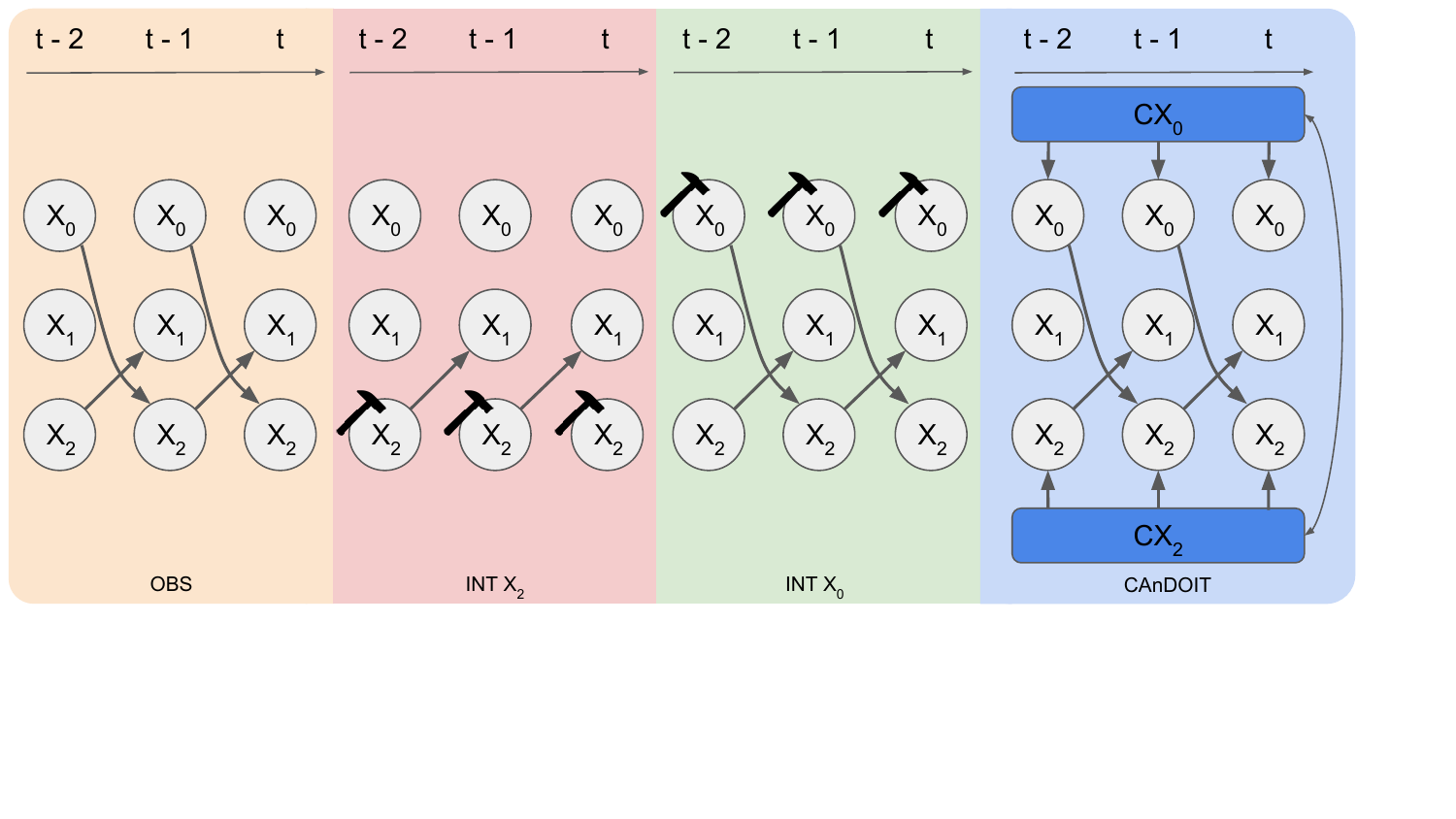}\caption{}\label{fig:approach_example_multi}
    \end{subfigure}
    \caption{CAnDOIT effectively employs context variables to handle observational and interventional data, resulting in a unified causal structure (right) that accommodates both types of data. In contrast, analysing these data separately leads to different causal structures for observations (left) and interventions (center).}\label{fig:approach_example}
\end{figure}

At this point, we can further clarify the concept of context nodes. Essentially, the context node is a dummy exogenous variable (i.e. a “meta-variable” that does not exist in the real system) that is used to inject the interventional data into the intervention variable without ignoring its parents. Since the context node is exogenous (by the JCI1 assumption) and models the intervention, the model’s structure does not change when transitioning between observational and interventional cases.
Following the JCI framework for causal discovery with both observational and interventional data,\mycite{mooij2020joint} the interventional process in CAnDOIT is generated by creating context nodes (e.g. $C\!X_{k}$) that are added as parent of the system variables (e.g. $X_k$) and govern their possible values. The context node affects its system variable instantaneously (at the same time step) injecting the interventional data into it and maintaining its value constant for the duration of the intervention. Note that, as the context node does not carry temporal information, i.e., its value does not change over time, we modelled it as a unique node in the graph confounding its corresponding system variable at all the time intervals (see \textbf{Figure~\ref{fig:approach_example}} CAnDOIT blocks). This way of modeling context node as parameter rather than a variable has been previously adopted in the literature.\mycite{kocaoglu2019characterization,yang2018characterizing}

\textbf{Figure~\ref{fig:approach_example_single}} shows an example of a context variable to handle a hard intervention. Initially, a causal structure is obtained using only observational data Figure~\ref{fig:approach_example_single}~(OBS). When intervening on $X_2$, all its input dependencies need to be removed, resulting in a different causal structure Figure~\ref{fig:approach_example_single}~(INT $X_2$). CAnDOIT instead creates a context node $C\!X_2$ connected to $X_2$ at each time step. Since the intervention is now assigned to the context variable $C\!X_2$, and the latter is not affected by any other variables (JCI1 assumption), there is no need to modify the original causal structure. As a result, we obtain a unified causal structure Figure~\ref{fig:approach_example_single}~(CAnDOIT) that represents both observational and interventional data. 
\textbf{Figure~\ref{fig:approach_example_multi}} demonstrates the application of CAnDOIT with multiple interventions. Figure~\ref{fig:approach_example_multi}~(OBS) shows the causal structure describing the observational case. Figure~\ref{fig:approach_example_multi}~(INT $X_2$)~and~(INT $X_0$) depict the two different causal models corresponding to the two interventions, $X_2$ and $X_0$, respectively. Figure~\ref{fig:approach_example_multi}~(CAnDOIT) illustrates the unified causal structure obtained by CAnDOIT.

Notably, in the case of multiple interventions, the context nodes are all connected by bidirected links between each pair of context nodes (JCI3). An example of a common situation where context variables are employed to model interventions and the context distribution contains no conditional independencies is what the JCI authors refer to as a \emph{diagonal design}, i.e., only one intervention is active at a time.\mycite{mooij2020joint} Applying this concept to the time-series domain, we model multiple interventions in such a way that only one intervention is active within a specific time interval.

\subsection{Faithfulness Assumption} 
In the context of constraint-based causal discovery, the so-called \emph{faithfulness assumption} plays a crucial role in ensuring that the inferred causal relationships from observational data represents accurately the true underlying causal structure. Essentially, this assumption asserts that all conditional independence relationships in the data are encoded in the reconstructed causal graph. This means that if a variable is conditionally independent from another, given a set of variables in the data, then no direct edge should exist between them in the graph.\mycite{peters2017elements}
Like its predecessor LPCMCI, CAnDOIT is a constraint-based causal discovery method that requires the faithfulness assumption, even with the introduction of new context variables for modeling interventions. This is crucial for the correctness of the algorithm.
Since we combine both observational and interventional data, our output (time series PAG) must be faithful to the conditional independencies present in both types of data. To ensure this, we test for (conditional) independence using only the pooled dataset (observational and interventional) and never consider the two cases separately. Testing for independence using only part of the data might produce results that are unfaithful to the remaining data.

From the example in Figure~\ref{fig:approach_example_single}, we can see how CAnDOIT reconstructs the causal model using both observational and interventional data, consistently with the faithfulness assumption. As shown in in the Figure~\ref{fig:approach_example_single} (CAnDOIT), the meta-system $\mathcal{M}$ created by CAnDOIT includes three system variables $(X_0, X_1, X_2)$ and a context variable $C\!X_2$, which models the intervention on $X_2$.
Although the hard intervention on the variable $X_2$ overrides the relationships between $X_2$ and its parents $P\!a(X_2) = X_0$, it does not compromise the faithfulness of the joint distribution $P(C\!X_2, X_0, X_1, X_2)$ with respect to the joint causal graph. Indeed, even though $X_2$ and $P\!a(X_2) = X_0$ are independent in the intervention setting, they are still dependent in the observational one. Therefore, as long as we do not test for independences in the subset of interventional data separately, but restrict ourselves to testing independences only in the pooled data set that combines all contexts, the faithfulness assumption remains valid.\mycite{mooij2020joint}

\subsection{CAnDOIT Algorithm}\label{subsec:candoit}

\begin{figure}[t]
    \centering
    \includegraphics[trim={0cm 0cm 5cm 7cm}, clip, width=\textwidth]{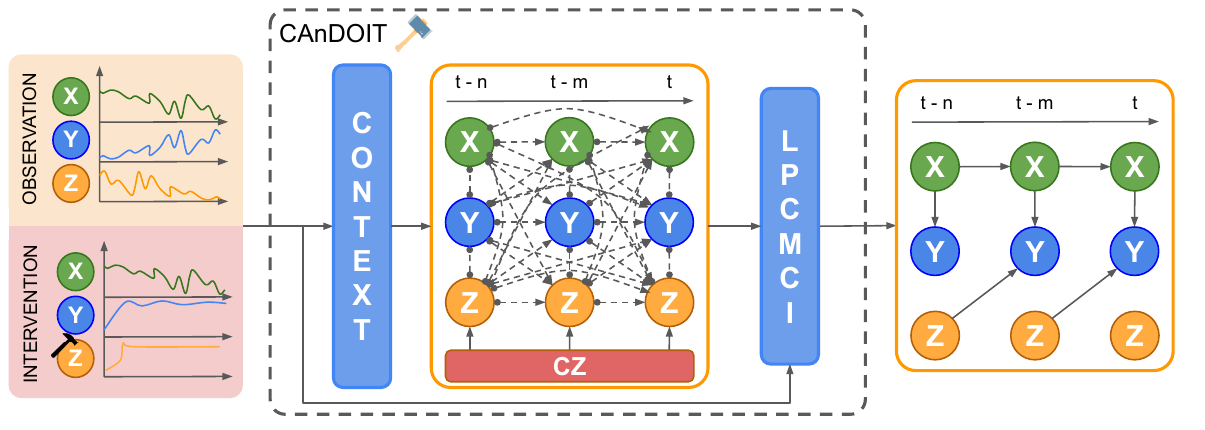}
    \caption{CAnDOIT’s block scheme representation. CAnDOIT processes observational and interventional data; the context block adds context variables ($C\!Z$) linked to the actual intervention variable ($Z$) with an instantaneous link ($C\!Z \rightarrow Z$); Finally, the LPCMCI block finalizes the causal discovery process.}
    \label{fig:approach_flowchart}
\end{figure}
\textbf{Figure~\ref{fig:approach_flowchart}} depicts a detailed flowchart of CAnDOIT, explaining each step of the algorithm with an example. In particular, the steps executed by our approach are as follows:
\begin{itemize}
    \item CAnDOIT takes observational and interventional data as input;
    \item Using the knowledge of the intervention target $Z$, the {\em context} block adds the context node $C\!Z$ to the set of variables considered in $\mathcal{M}$, plus an instantaneous link $C\!Z \rightarrow Z$ to the initial causal structure, i.e., a fully connected graph that is the starting point of the LPCMCI algorithm;
    \item The system variables $(X, Y, Z)$, along with the context node $C\!Z$ are injected into the causal discovery block;
    \item LPCMCI performs the causal analysis on the meta-system $\mathcal{M}$ and then removes both the context variable $C\!Z$ and the link $C\!Z \rightarrow Z$ before returning the causal model;
    \item CAnDOIT outputs a time-series PAG.
\end{itemize}
A detailed pseudo-code explanation of our approach is presented in \textbf{Algorithm~\ref{alg:approach_pseudocode}}. A Python implementation of CAnDOIT is also publicly available\footnote{\url{https://github.com/lcastri/causalflow}}.

\begin{algorithm}[H]
\small
\caption{CAnDOIT} \label{alg:approach_pseudocode}
\begin{algorithmic}[1]
\REQUIRE{
    obs. $D_{obs}$ and int. $D_{int}$ ts data, int. target variables $X_{i}$, 
    significance level $\alpha$,
    min and max time lag $\tau_{min}$, $\tau_{max}$.
    }
    \STATE $C\!M_0 \leftarrow$ fully connected PAG with $\circlearrow$ for lagged dependencies and $\circlecircle$ for contemporaneous dependencies \hspace{\fill}{\footnotesize$\leftarrow$ LPCMCI starting point}
    \STATE $\mathcal{M} \leftarrow \mathcal{I}$ add the set of system variables $X = (X_i)_{i\in\mathcal{I}}$ to the meta-system $\mathcal{M}$
    \STATE \textbf{for each} int. target variables $X_{i}$ \textbf{do}
        \STATE\ind $C\!X_{k} \leftarrow$ create the context variable $C\!X_{k}$ associated to the intervention system variable $X_{i}$
        \STATE\ind $\mathcal{M} \leftarrow C\!X_{k}$ add $C\!X_{k}$ to the meta-system $\mathcal{M}$
        \STATE\ind $C\!M_0 \leftarrow$ add $C\!X_{k}$ to the LPCMCI initial condition $C\!M_0$
        \STATE\ind \textbf{for each} $\tau$ in range($\tau_{min}$, $\tau_{max}$) \textbf{do}
            \STATE\ind\ind $C\!M_0 \leftarrow$ add the link $C\!X_{k} \rightarrow X_{i}(t-\tau)$ to $C\!M_0$
\STATE $D_s \leftarrow$ $[D_{obs}, D_{int}]$
\STATE $C\!M$ = \texttt{LPCMCI}($D_s$, $\alpha$, $\tau_{min}$, $\tau_{max}$, $C\!M_0$)
\STATE $C\!M$ $\leftarrow$ remove context variables $CX_{k}$ and related links
\RETURN $C\!M$ \hspace{\fill}{\footnotesize$\leftarrow$ time-series PAG}
\end{algorithmic}
\end{algorithm}

Being based on LPCMCI, our CAnDOIT inherits its necessary conditions for proper functioning: Causal Markov Condition, Faithfulness, Acyclicity.
Furthermore, like its predecessor, CAnDOIT can adapt to any type of data, including linear and nonlinear relationships, multiple time lags, various types of noise, and it cannot detect cyclical relationships. It retains the output format of a time-series Partial Ancestral Graph (PAG). Specifically, CAnDOIT produces a time-series PAG with a number of layers determined by the algorithm parameters $\tau_{min}$ and $\tau_{max}$ (see Algorithm~\ref{alg:approach_pseudocode} inputs). By default, $\tau_{min}$ is set to 0 to account for the instantaneous links created for the context variables. On the other hand, $\tau_{max}$ represents the maximum time delay considered when the algorithm performs conditional independence tests between variables across different time steps. Consequently, the time-series PAG consists of $\tau_{max} + 1$ layers, corresponding to the time steps $t - \tau_{max}, t - (\tau_{max} - 1), \ldots, t$.

PAGs are used to represent the Markov equivalence class of Maximal Ancestral Graphs (MAGs). The latter extend the DAGs representation by including also the bidirected link ($\leftrightarrow$) to represent variables confounded by a latent confounder. PAGs further generalize MAGs by incorporating additional edge types, specifically $\circlearrow$ and $\circlecircle$, to handle uncertainties in edge orientations.
For example, in a PAG, a link $X \circlearrow Y$ corresponds to two possible MAGs: $X \rightarrow Y$ (where $X$ is an ancestor of $Y$) or $X \leftrightarrow Y$ (where $X$ and $Y$ are confounded by a latent variable). Similarly, a link $X \circlecircle Y$ in a PAG represents two possible MAGs: $X \rightarrow Y$ (where $X$ is an ancestor of $Y$) or $Y \rightarrow X$ (where $Y$ is an ancestor of $X$).

\section{Evaluation on Random Synthetic Models}\label{sec:randomdag}
We designed experiments to assess the accuracy and effectiveness of our approach in handling interventional data and to examine its impact on the causal discovery process. These experiments aimed to evaluate our method's performance in various scenarios: linear and nonlinear systems, with and without latent confounders, and involving single or multiple interventions.

\subsection{Random Synthetic Models}\label{subsec:toy_problem}
To evaluate our approach's effectiveness in handling interventional data and its impact on the causal structure, we devised five testing strategies, denoted as $S_1$, $S_2$, $S_3$, $S_4$, and $S_5$. 
In the first strategy ($S_1$), we assessed our approach’s performance with linear systems while varying the number of observable variables and without hidden confounders. Strategy $S_2$ extends $S_1$ by introducing hidden confounders while retaining the system as linear and maintaining the same range of the number of variables. In both $S_1$ and $S_2$, only a single intervention is conducted.
In strategy $S_3$, we evaluated how CAnDOIT performs with linear systems and hidden confounders when multiple interventions are applied, while keeping the number of observable variables fixed. Strategies $S_4$ and $S_5$ mirror $S_2$ and $S_3$, respectively, but focus on nonlinear systems.

\begin{figure}
    \centering
    \begin{subfigure}{.33\textwidth}
        \includegraphics[trim={0cm 0cm 16cm 0cm}, clip, width=\textwidth]{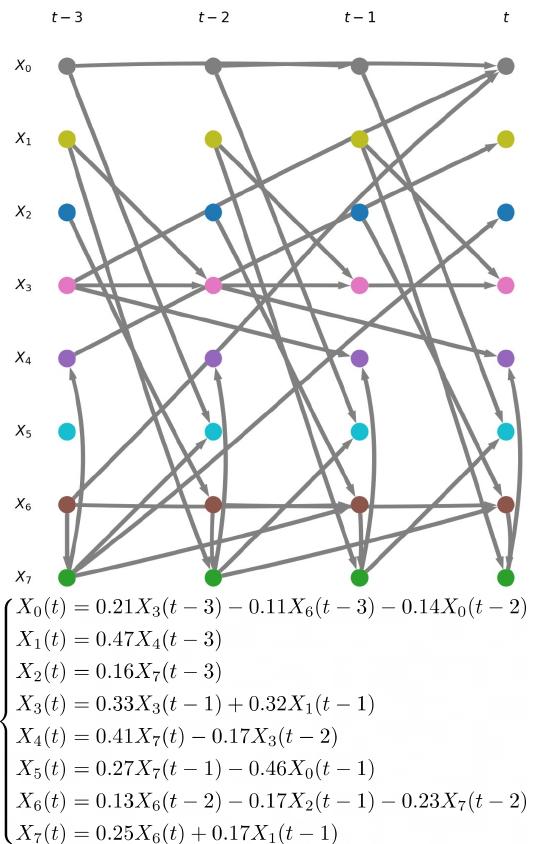}\caption{}\label{fig:s1_randoomgraph}
    \end{subfigure}
    \begin{subfigure}{.27\textwidth}
        \includegraphics[trim={0cm 0cm 17.5cm 0cm}, clip, width=\textwidth]{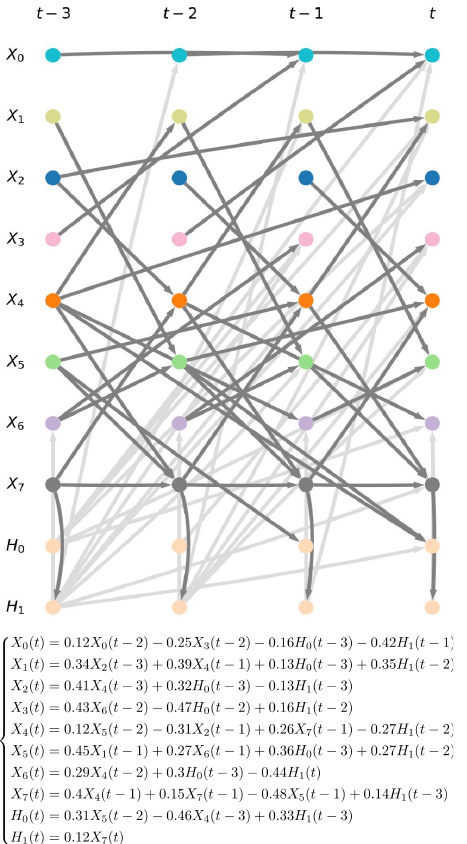}\caption{}\label{fig:s2_randoomgraph}
    \end{subfigure}
    \begin{subfigure}{.39\textwidth}
        \includegraphics[trim={0cm 0cm 14cm 0cm}, clip, width=\textwidth]{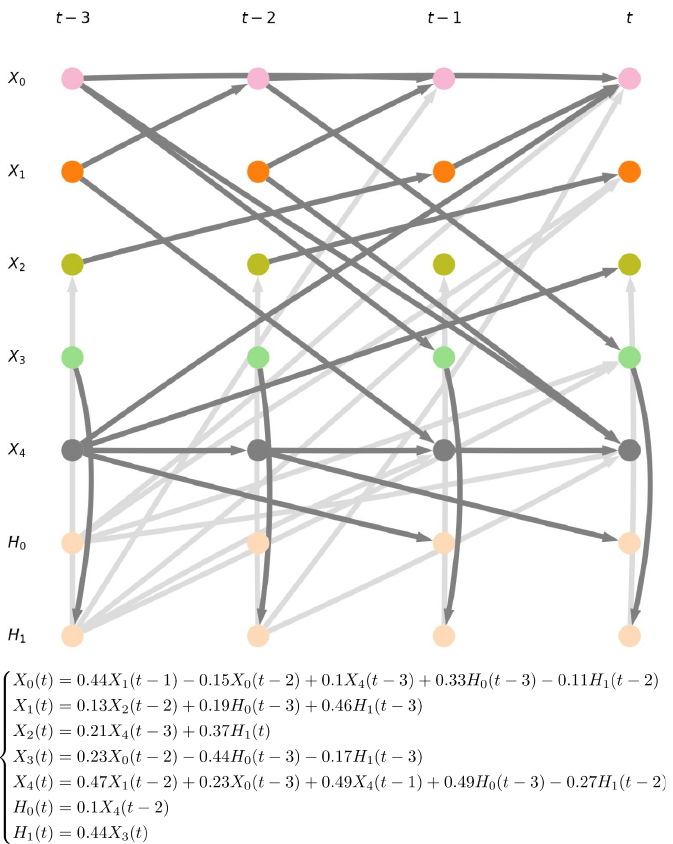}\caption{}\label{fig:s3_randoomgraph}
    \end{subfigure}\\
    \begin{subfigure}{.35\textwidth}
        \includegraphics[trim={0cm 0cm 17cm 0cm}, clip, width=\textwidth]{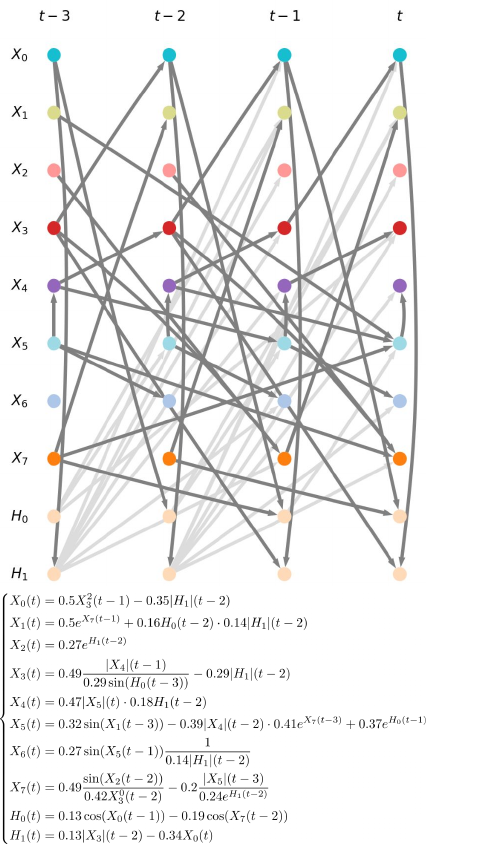}\caption{}\label{fig:s4_randoomgraph}
    \end{subfigure}
    \begin{subfigure}{.45\textwidth}
        \includegraphics[trim={0cm 0cm 15cm 0cm}, clip, width=\textwidth]{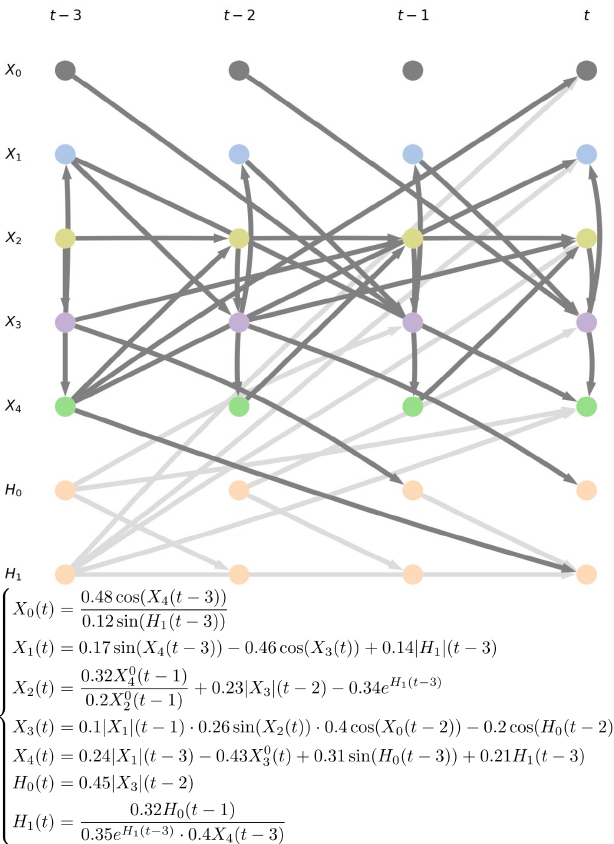}\caption{}\label{fig:s5_randoomgraph}
    \end{subfigure}
    \caption{  
    Causal models randomly generated with their corresponding systems of equations on the bottom.
    (a) An example of a linear system for the $S_1$ evaluation strategy, which has no hidden confounders;
    (b) A random linear system for the $S_2$ evaluation strategy, including two hidden confounders ($H_0,~H_1$);
    (c) An example of a five-variable random linear system with two hidden confounders ($H_0,~H_1$) for the $S_3$ evaluation strategy.
    (d) and (e) present nonlinear counterparts of (b) and (c), respectively.
    For ease of reading, we present only examples with a maximum time lag of 3, 8 observable variables, and 2 hidden confounders. However, in the $S_1,~S_2,~S_4$ evaluation strategies, the maximum number of variables was 12, and the maximum number of hidden confounders was 3.
    }
    \label{fig:example_randomdag}
\end{figure}

To facilitate the aforementioned evaluations, we developed a synthetic model generator capable of creating random systems of equations with hidden confounders. This tool offers various adjustable parameters, including time-series' length, number of observable variables, observable parents per variable (link density), hidden confounders, and confounded variables per hidden confounder. Moreover, it includes also noise configuration, minimum and maximum time delay to consider in the equations, coefficient range, plus functional forms and operators used to link various equation's terms. 
With this generator, we can create ground-truth causal models to test our algorithm and generate observational and interventional data based on the generated causal structure. This enables us to simulate different scenarios and thoroughly analyze the behavior of our approach under various conditions. A detailed explanation of the random-model generator is presented in Appendix~\ref{appx:RandomDAG_setting}. Examples of randomly-generated causal models for each specific evaluation strategy are shown in \textbf{Figure~\ref{fig:example_randomdag}}.

Some model generators and causal discovery benchmarks have been introduced in the literature, for example CauseMe,\mycite{runge_detecting_2019} a collection of synthetic, hybrid, and real observational data mostly for climate and weather scenarios. Other random-model generators have been proposed\mycite{dgp_causal_discovery_time_series}, however none of them can generate entirely random causal structures along with both observational and interventional data like the one here developed.

Our synthetic model evaluation is crucial to validate CAnDOIT's performance in retrieving more accurate causal models compared to other methods. Opportunely tuned, our random-model generator is not constrained by any assumptions regarding model linearity, effectively excluding from the evaluation noise-based discovery methods.\mycite{peters2013causal,hyvarinen2010estimation} 
As CAnDOIT is based on the LPCMCI causal discovery algorithm, the latter was used as the benchmark for this evaluation.

\subsection{Evaluation Setting}\label{subsec:toy_evaluation}
We evaluated CAnDOIT using the random-model generator mentioned in Section~\ref{subsec:toy_problem}.
Five evaluation strategies were devised. For each strategy, we first conducted causal discovery analysis with LPCMCI. Based on its result, we then selected which variable(s) to intervene on. The choice of intervention variable(s) was determined by the observable variable(s) identified by LPCMCI as having ambiguous links ($\circlearrow$ and $\circlecircle$). The goal of these interventions was to (partially) resolve the ambiguities in these links. 
In cases where LPCMCI returned a causal model with ambiguous links originating from multiple nodes, we performed an intervention for each of these nodes. In the evaluation results presented in Section~\ref{subsec:toy_results}, we report two curves for each metric, both describing CAnDOIT's performance: \emph{CAnDOIT\_mean}, which depicts the average for the analyzed metric across all interventions, and \emph{CAnDOIT\_best}, which shows the result that produced the most accurate (highest $F_1$-Score) causal model.

The $S_1$ evaluation strategy analyzes how CAnDOIT behaves with linear systems, where the number of observable variables ranges from 5 to 12 and there are no latent confounders. $S_2$ extends $S_1$ by introducing a random number of latent confounders (from 1 to 3) that affect a specific number of observable variables. In both $S_1$ and $S_2$, we perform a single intervention following the strategy explained above.
%
In the $S_3$ evaluation strategy, we analyze how CAnDOIT performs with an increasing number of interventions while keeping the number of observable variables fixed. We focus again on linear systems, this time composed of 5 observable variables plus a random number of hidden confounders (from 1 to 3). The number of interventions ranges from 1 to 3.
%
As mentioned earlier, the $S_4$ and $S_5$ evaluation strategies are the nonlinear counterparts of $S_2$ and $S_3$, respectively.

The two algorithms in the comparison -- LPCMCI and CAnDOIT -- used observational time-series data. For CAnDOIT, in case of $S_1,~S_2,~S_4$, we also conducted a single intervention on the variable originating ambiguous links, while for $S_3$ and $S_5$, we conducted up to three interventions. To manage the computational cost and ensure a fair comparison, both algorithms receive the same amount of data. Specifically, LPCMCI used 1300 observational data samples, while CAnDOIT used 1000 samples of observational data and 300 samples of interventional data. The 300 samples of interventional data were equally divided among the number of interventions. Moreover, the minimum and the maximum time lags considered in the causal analysis were set to 0 and 3, respectively. Finally, the link density parameter, which corresponds to the number of observable parent per variable, was set to 3.

Inspired by the evaluation analyses done in the literature for PCMCI, PCMCI\textsuperscript{+} and LPCMCI,\mycite{runge_causal_2018,runge2020discovering,gerhardus2020high} we used similar noise configurations (uniform, Gaussian, Weibull) and the same value range for the coefficients that interconnect different terms within the equation, i.e., $[0.1, 0.5]$. This choice of parameters allows to create random systems without encountering divergences or singularities, which commonly arise when attempting to generate time-series equations, especially a broad range of parameter values and autocorrelation terms is considered. Moreover, differently from PCMCI\textsuperscript{+} and LPCMCI's evaluation strategies, which used a predetermined non-linear functional form, we employed several functional forms, both linear and non-linear, randomly chosen. This approach increased the sparsity and diversified the experiments to explore a broader range of scenarios. The complete set of parameters used in our evaluation strategies is listed and discussed in detail in \textbf{Tables~\ref{tab:linear_experimental_settings}}~and~\textbf{\ref{tab:nonlinear_experimental_settings}} of Appendix~\ref{appx:RandomDAG_setting}.

\subsection{Evaluation Metrics} 
The evaluation metrics were categorized into two main groups. The first one comprises metrics to assess CAnDOIT's ability in removing ambiguous links, and includes mean False Positive Rate~(FPR), mean number of ambiguous links~(Uncertainty), and mean number of equivalent MAGs that the PAG output by the algorithms can represent (PAG Size) across all tests. The second category instead evaluates CAnDOIT's overall performance in recovering the correct causal model and its execution time. These last metrics include mean Structural Hamming Distance~(SHD), mean $F_1$-Score, and mean execution time across all tests. All the means were computed based on 25 test runs with different random systems for each configuration.
Note that for the FPR, SHD, and $F_1$-Score metrics, we adopted the same approach used in LPCMCI's evaluation metrics.\mycite{gerhardus2020high} Specifically, we not only considered the existence of a link between two nodes but also the tail and head markers of the links. Thus, in calculating these metrics, we measured both the adjacency (presence of a specific link) and the orientation (tail and head markers of a specific link).

For sake of clarity, since less known in the literature, we explain Uncertainty and PAG size metrics a little more in details.
The Uncertainty metric quantifies the number of ambiguous links present in the discovered causal model ($\circlearrow$ and $\circlecircle$). This metric is utilized to assess CAnDOIT's capability in removing ambiguities. Consequently, a lower value indicates better performance.
The PAG Size metric quantifies the number of MAGs equivalent to the discovered one. It calculates this number based on the count of ambiguous links in the estimated causal model. As already explained in Section~\ref{subsec:candoit}, each ambiguous link can represent two possible links: $\circlearrow$ can represent either $\rightarrow$ or $\leftrightarrow$, while $\circlecircle$ can represent either $\rightarrow$ or $\leftarrow$. Consequently, for each ambiguous link present in the final causal model, two equivalent MAGs are generated. Finally, the PAG Size metric is defined as follows:
\begin{equation}
    \text{PAG Size} = 2^{\text{Uncertainty}}
\end{equation}
This metric is employed to evaluate CAnDOIT's effectiveness in enhancing identifiability by using interventional data.


\subsection{Experimental Results on Synthetic Models}\label{subsec:toy_results}
\begin{figure}
    \centering
    \begin{subfigure}{.475\textwidth}
        \includegraphics[trim={0.15cm 0cm 1.2cm 0cm}, clip, width=\textwidth]{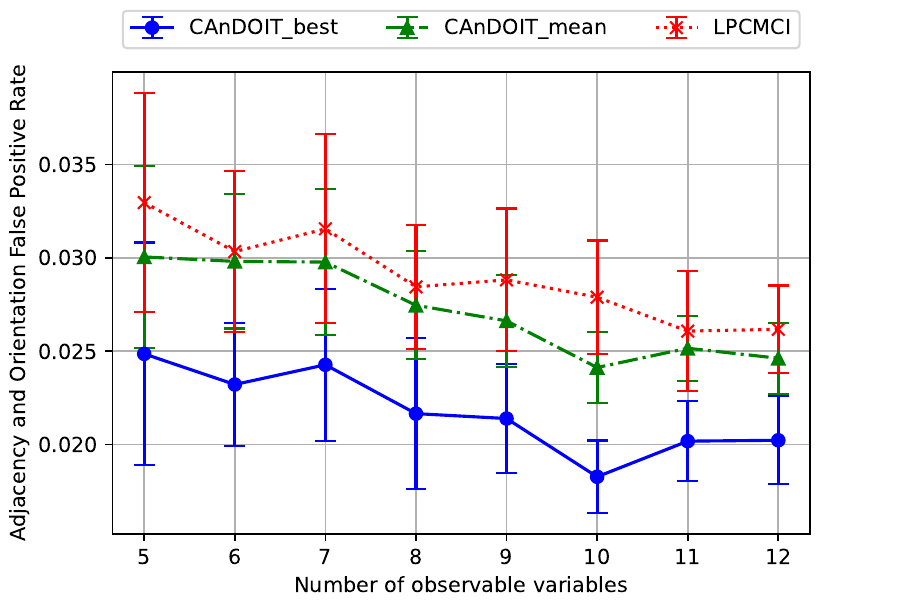}\caption{}\label{fig:s1_fpr}
    \end{subfigure}
    \qquad
    \begin{subfigure}{.475\textwidth}
        \includegraphics[trim={0.35cm 0cm 1.2cm 0cm}, clip, width=\textwidth]{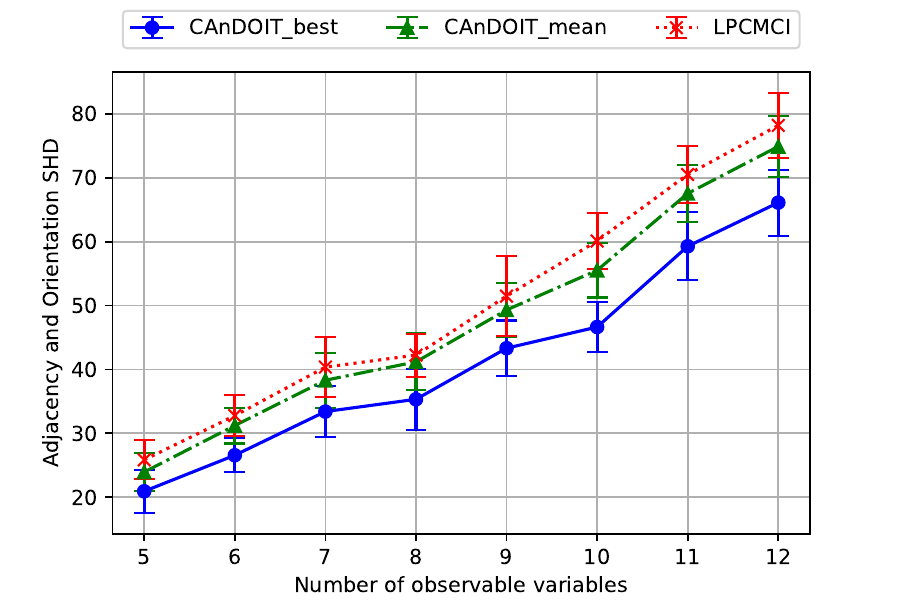}\caption{}\label{fig:s1_shd}
    \end{subfigure}\\
    \begin{subfigure}{.475\textwidth}
        \includegraphics[trim={0.35cm 0cm 1.2cm 1.2cm}, clip, width=\textwidth]{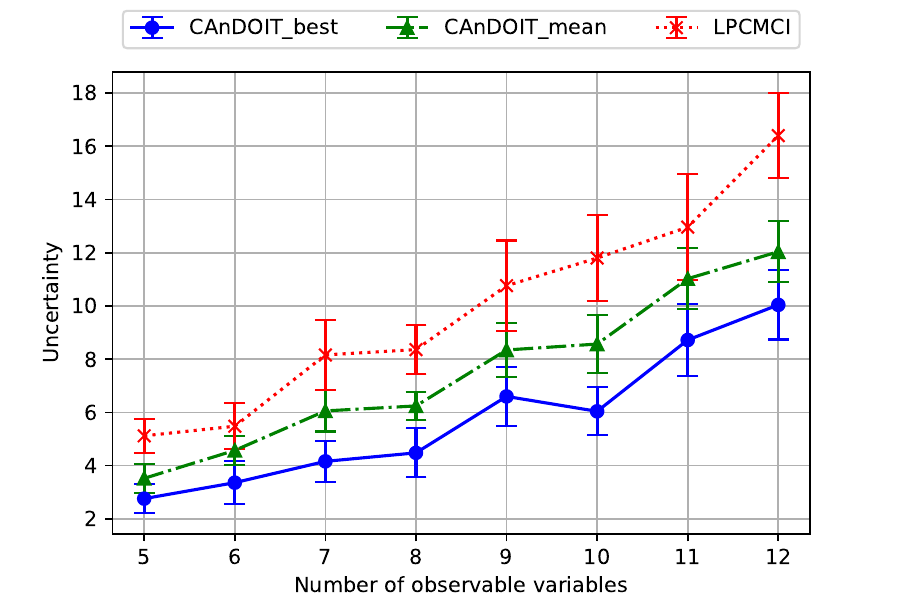}\caption{}\label{fig:s1_uncertainty}
    \end{subfigure}
    \qquad
    \begin{subfigure}{.475\textwidth}
        \includegraphics[trim={0.35cm 0cm 1.2cm 1.1cm}, clip, width=\textwidth]{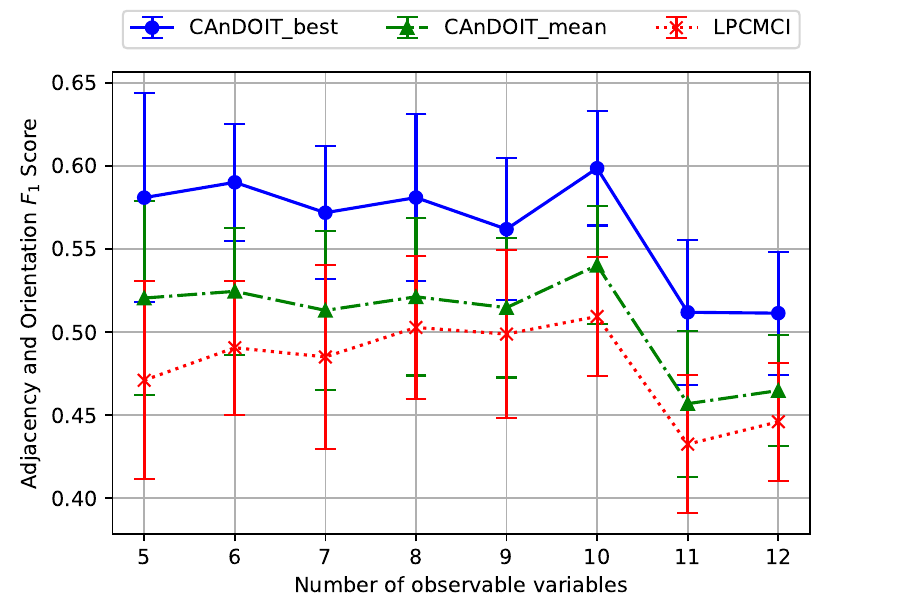}\caption{}\label{fig:s1_f1}
    \end{subfigure}\\
    \begin{subfigure}{.475\textwidth}
        \includegraphics[trim={0.35cm 0cm 1.2cm 1.2cm}, clip, width=\textwidth]{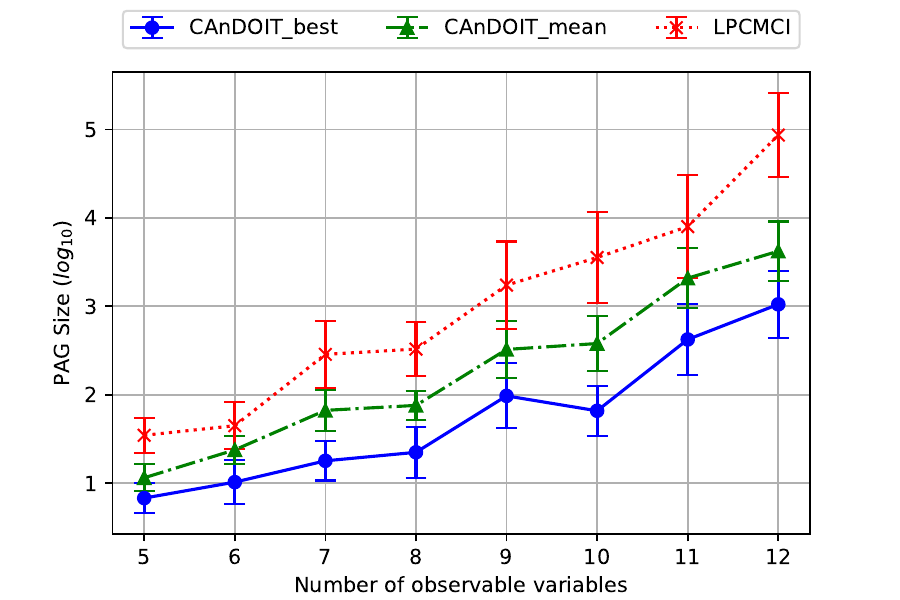}\caption{}\label{fig:s1_pag}
    \end{subfigure}
    \qquad
    \begin{subfigure}{.475\textwidth}
        \includegraphics[trim={0.35cm 0cm 1.2cm 1.1cm}, clip, width=\textwidth]{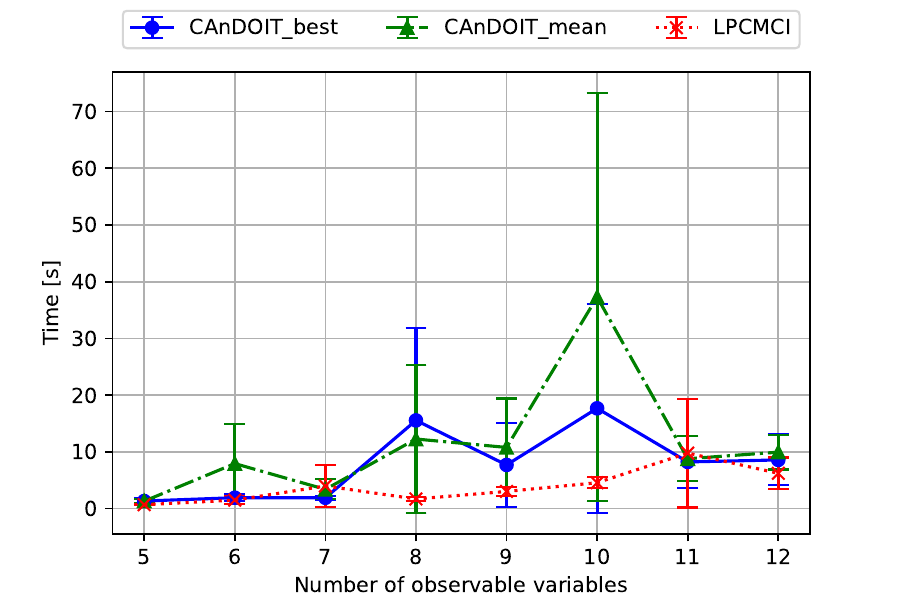}\caption{}\label{fig:s1_time}
    \end{subfigure}\\
    \caption{LPCMCI (red dotted line), CAnDOIT\_mean (green dashed line) and CAnDOIT\_best (blue) in $S_1$ analysis: linear systems with a number of observable variables ranging from 5 to 12 and no hidden confounders. (a) False Positive Rate (FPR); (b) Structural Hamming Distance (SHD); (c) Uncertainty; (d) $F_1$-Score; (e) PAG Size (reported in logarithmic scale); (f) Time (expressed in seconds).}
    \label{fig:exp_S1}
\end{figure}

\begin{figure}
    \centering
    \begin{subfigure}{.475\textwidth}
        \includegraphics[trim={0.15cm 0cm 1.2cm 0cm}, clip, width=\textwidth]{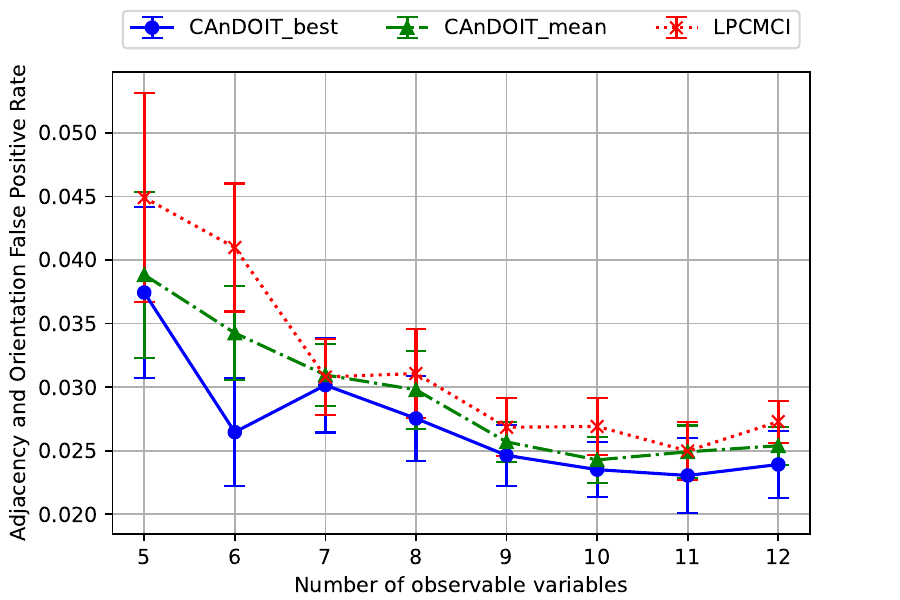}\caption{}\label{fig:s2_fpr}
    \end{subfigure}
    \qquad
    \begin{subfigure}{.475\textwidth}
        \includegraphics[trim={0.35cm 0cm 1.2cm 0cm}, clip, width=\textwidth]{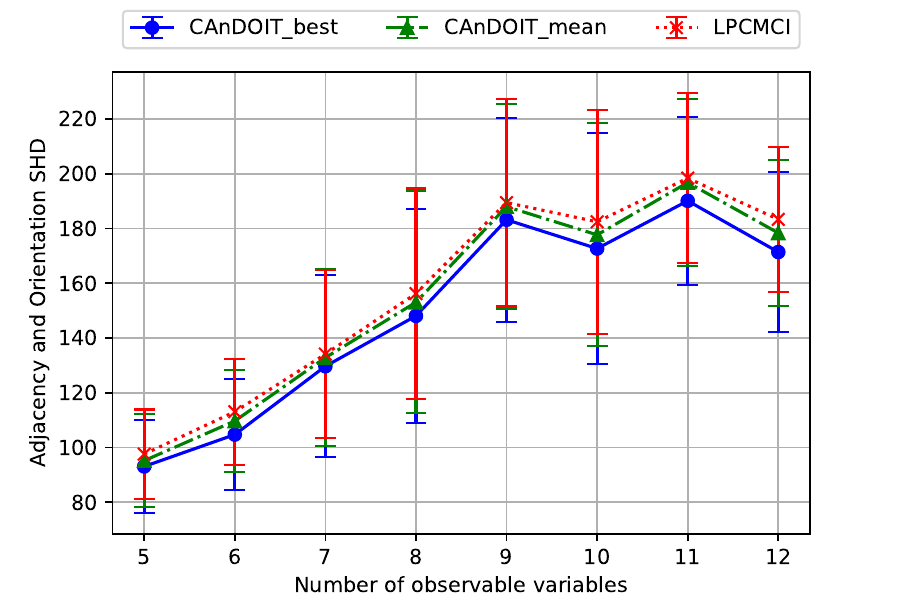}\caption{}\label{fig:s2_shd}
    \end{subfigure}\\
    \begin{subfigure}{.475\textwidth}
        \includegraphics[trim={0.35cm 0cm 1.2cm 1.2cm}, clip, width=\textwidth]{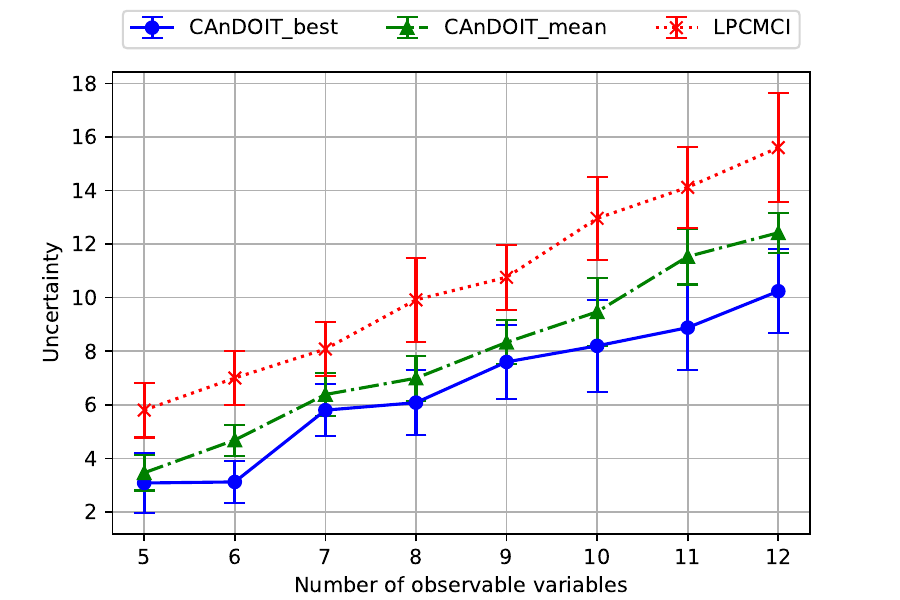}\caption{}\label{fig:s2_uncertainty}
    \end{subfigure}
    \qquad
    \begin{subfigure}{.475\textwidth}
        \includegraphics[trim={0.1cm 0cm 1.2cm 1.1cm}, clip, width=\textwidth]{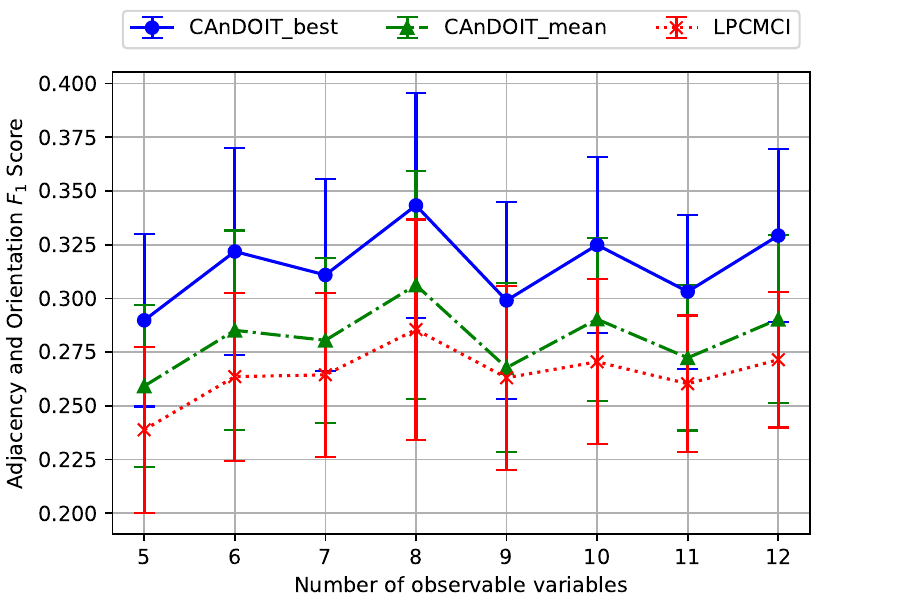}\caption{}\label{fig:s2_f1}
    \end{subfigure}\\
    \begin{subfigure}{.475\textwidth}
        \includegraphics[trim={0.35cm 0cm 1.2cm 1.2cm}, clip, width=\textwidth]{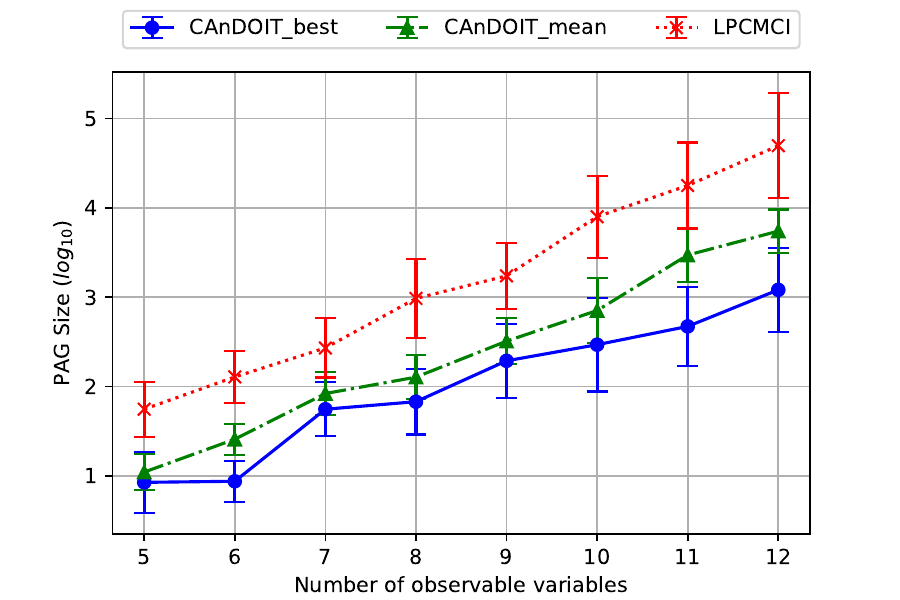}\caption{}\label{fig:s2_pag}
    \end{subfigure}
    \qquad
    \begin{subfigure}{.475\textwidth}
        \includegraphics[trim={0.35cm 0cm 1.2cm 1.1cm}, clip, width=\textwidth]{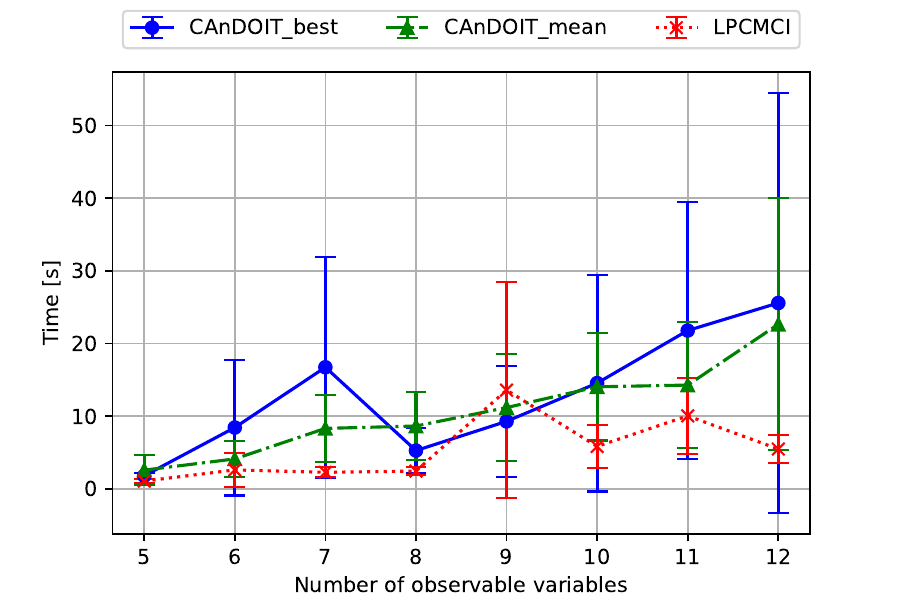}\caption{}\label{fig:s2_time}
    \end{subfigure}\\
    \caption{LPCMCI (red dotted line), CAnDOIT\_mean (green dashed line) and CAnDOIT\_best (blue) in $S_2$ analysis: linear systems with a number of observable variables ranging from 5 to 12 and a random  number of hidden confounders (from 1 to 3). (a) False Positive Rate (FPR); (b) Structural Hamming Distance (SHD); (c) Uncertainty; (d) $F_1$-Score; (e) PAG Size (reported in logarithmic scale); (f) Time (expressed in seconds).}
    \label{fig:exp_S2}
\end{figure}

\begin{figure}
    \centering
    \begin{subfigure}{.475\textwidth}
        \includegraphics[trim={0.15cm 0cm 1.2cm 0cm}, clip, width=\textwidth]{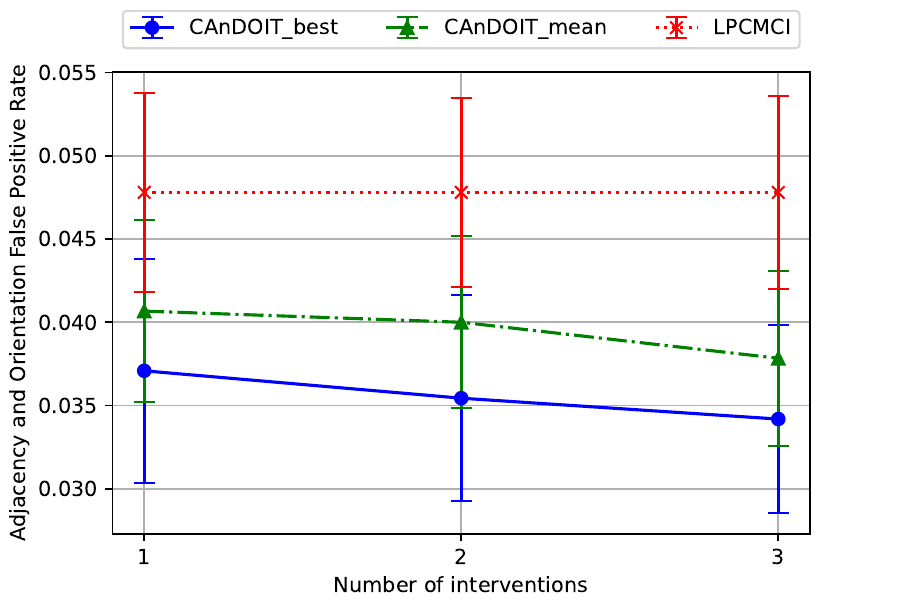}\caption{}\label{fig:s3_fpr}
    \end{subfigure}
    \qquad
    \begin{subfigure}{.475\textwidth}
        \includegraphics[trim={0.35cm 0cm 1.2cm 0cm}, clip, width=\textwidth]{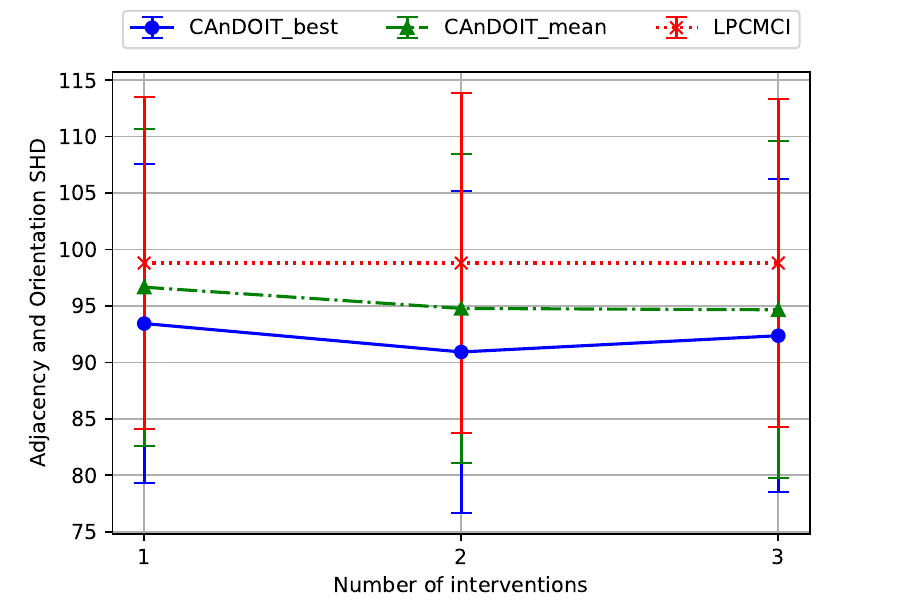}\caption{}\label{fig:s3_shd}
    \end{subfigure}\\
    \begin{subfigure}{.475\textwidth}
        \includegraphics[trim={0.35cm 0cm 1.2cm 1.2cm}, clip, width=\textwidth]{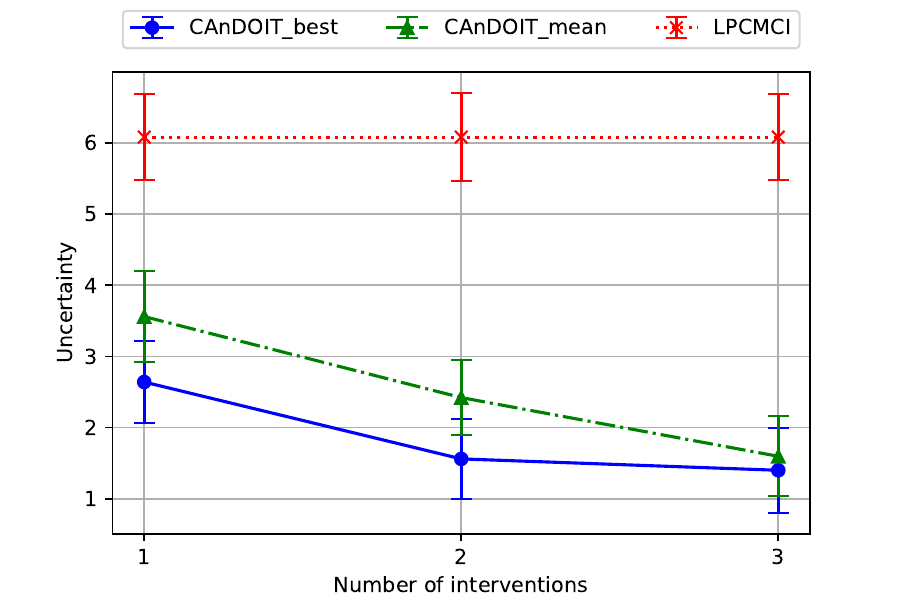}\caption{}\label{fig:s3_uncertainty}
    \end{subfigure}
    \qquad
    \begin{subfigure}{.475\textwidth}
        \includegraphics[trim={0.35cm 0cm 1.2cm 1.1cm}, clip, width=\textwidth]{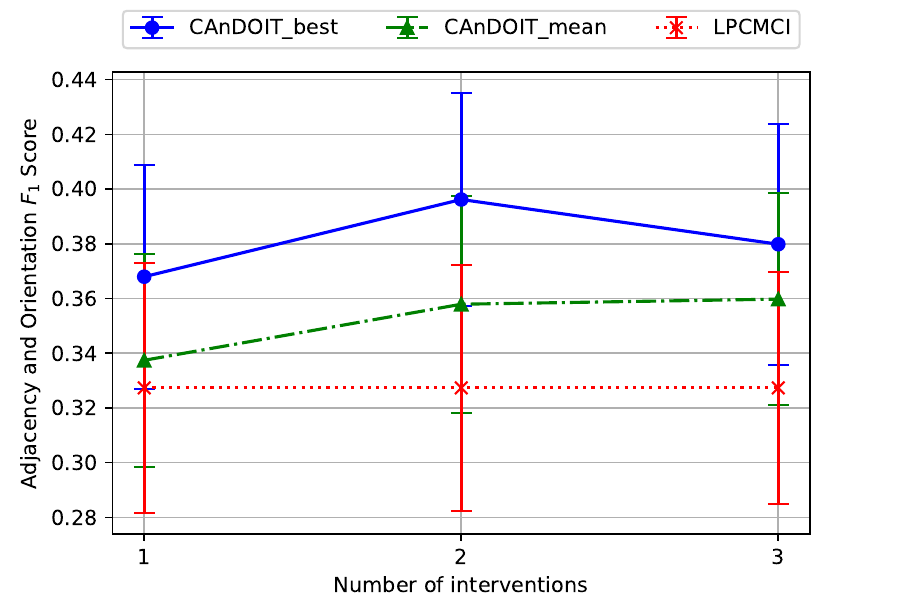}\caption{}\label{fig:s3_f1}
    \end{subfigure}\\
    \begin{subfigure}{.475\textwidth}
        \includegraphics[trim={0.35cm 0cm 1.2cm 1.2cm}, clip, width=\textwidth]{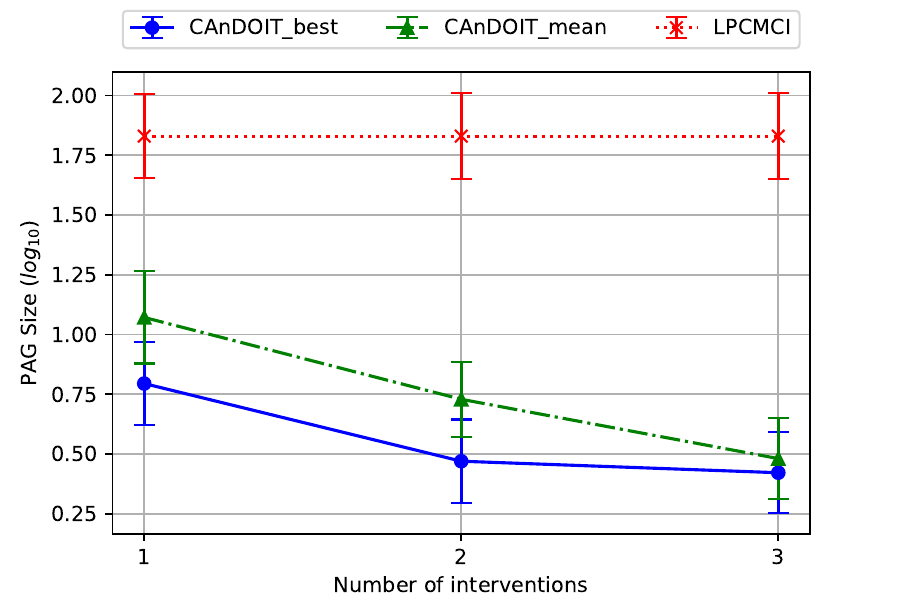}\caption{}\label{fig:s3_pag}
    \end{subfigure}
    \qquad
    \begin{subfigure}{.475\textwidth}
        \includegraphics[trim={0.35cm 0cm 1.2cm 1.1cm}, clip, width=\textwidth]{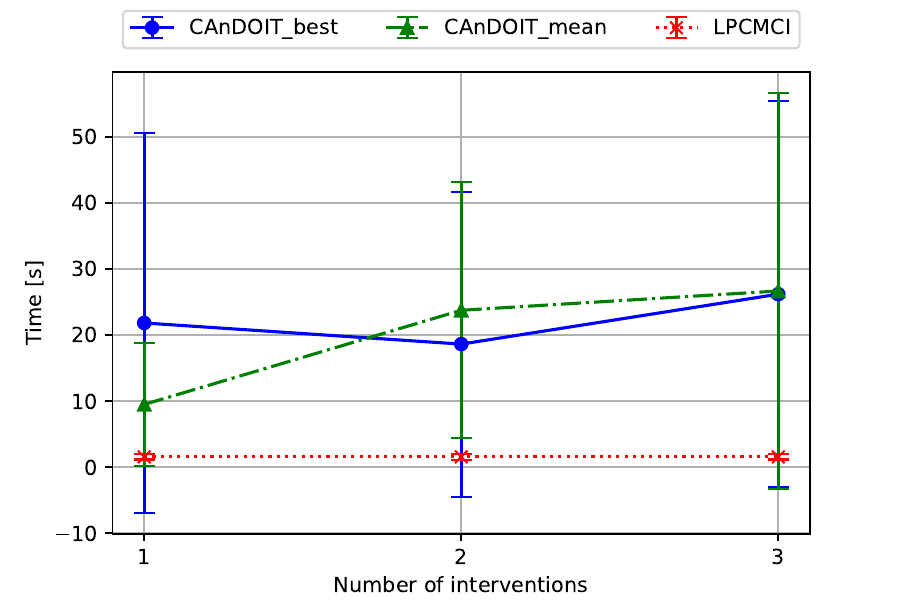}\caption{}\label{fig:s3_time}
    \end{subfigure}\\
    \caption{LPCMCI (red dotted line), CAnDOIT\_mean (green dashed line) and CAnDOIT\_best (blue) in $S_3$ analysis: linear systems with 5 observable variables, a random  number of hidden confounders (from 1 to 3) and an increasing number of interventions, ranging from 1 to 3. (a) False Positive Rate (FPR); (b) Structural Hamming Distance (SHD); (c) Uncertainty; (d) $F_1$-Score; (e) PAG Size (reported in logarithmic scale); (f) Time (expressed in seconds).}
    \label{fig:exp_S3}
\end{figure}

\begin{figure}
    \centering
    \begin{subfigure}{.475\textwidth}
        \includegraphics[trim={0.15cm 0cm 1.2cm 0cm}, clip, width=\textwidth]{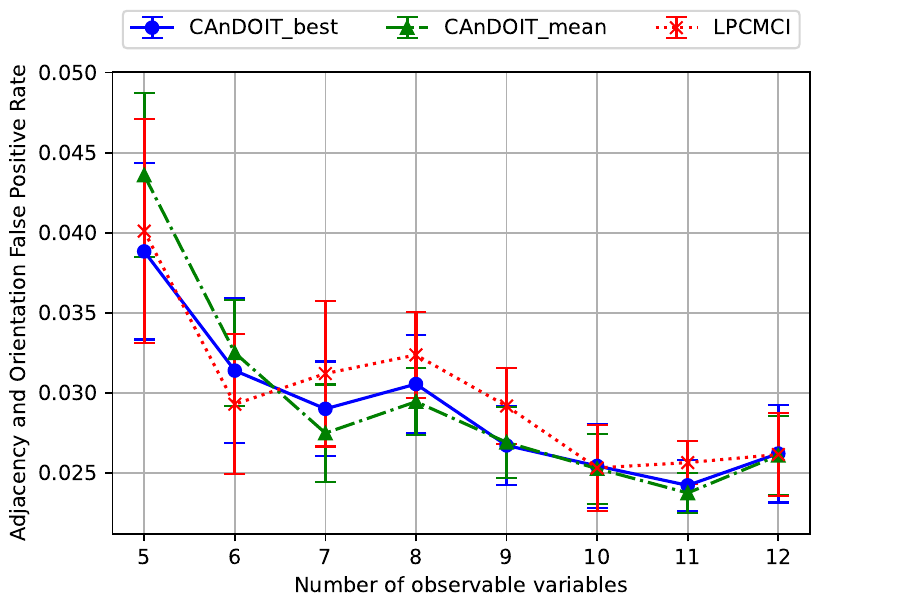}\caption{}\label{fig:s4_fpr}
    \end{subfigure}
    \qquad
    \begin{subfigure}{.475\textwidth}
        \includegraphics[trim={0.35cm 0cm 1.2cm 0cm}, clip, width=\textwidth]{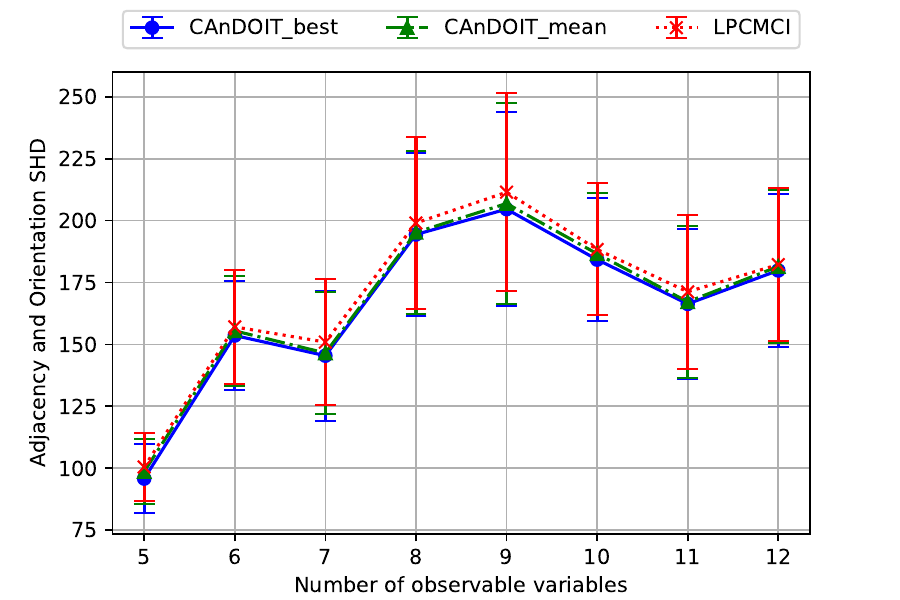}\caption{}\label{fig:s4_shd}
    \end{subfigure}\\
    \begin{subfigure}{.475\textwidth}
        \includegraphics[trim={0.35cm 0cm 1.2cm 1.2cm}, clip, width=\textwidth]{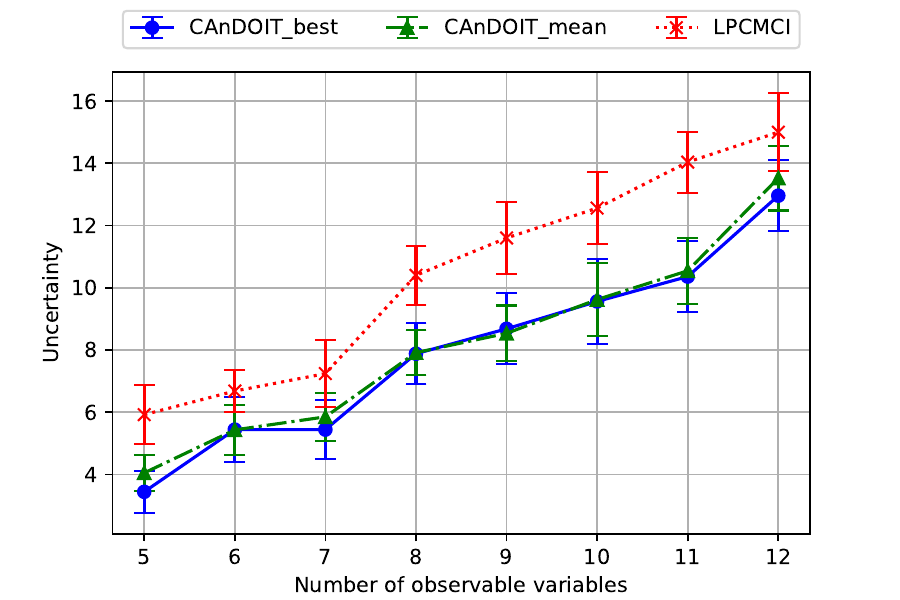}\caption{}\label{fig:s4_uncertainty}
    \end{subfigure}
    \qquad
    \begin{subfigure}{.475\textwidth}
        \includegraphics[trim={0.1cm 0cm 1.2cm 1.1cm}, clip, width=\textwidth]{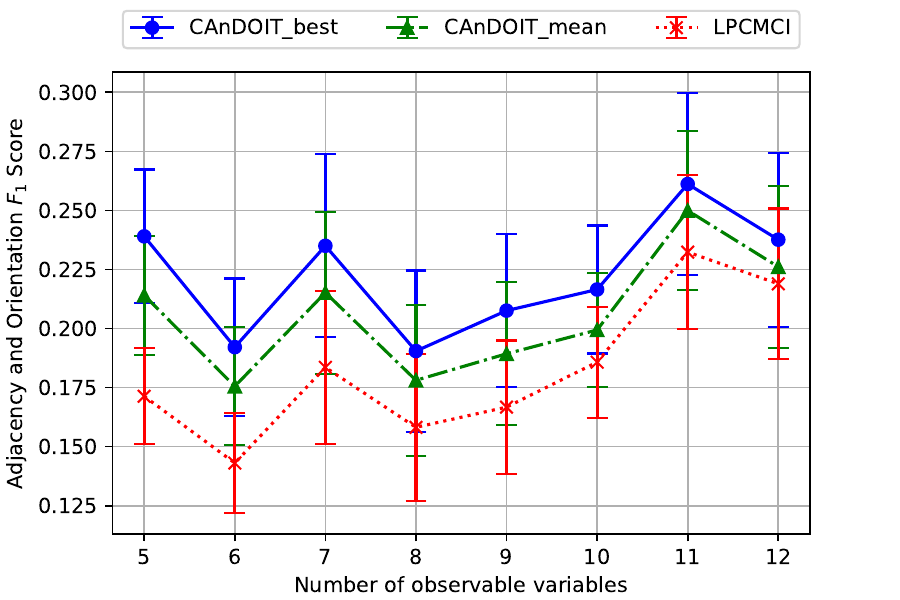}\caption{}\label{fig:s4_f1}
    \end{subfigure}\\
    \begin{subfigure}{.475\textwidth}
        \includegraphics[trim={0.35cm 0cm 1.2cm 1.2cm}, clip, width=\textwidth]{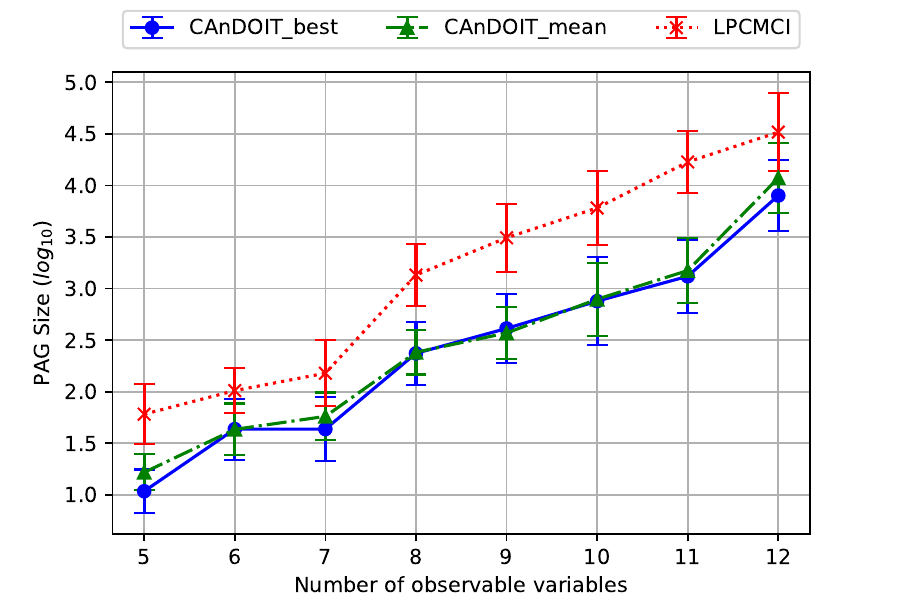}\caption{}\label{fig:s4_pag}
    \end{subfigure}
    \qquad
    \begin{subfigure}{.475\textwidth}
        \includegraphics[trim={0.35cm 0cm 1.2cm 1.1cm}, clip, width=\textwidth]{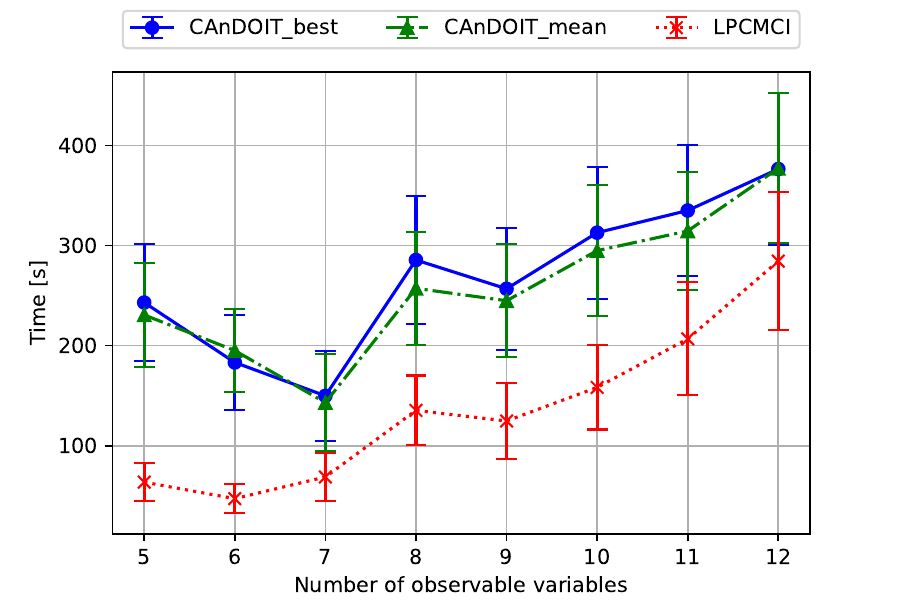}\caption{}\label{fig:s4_time}
    \end{subfigure}\\
    \caption{LPCMCI (red dotted line), CAnDOIT\_mean (green dashed line) and CAnDOIT\_best (blue) in $S_4$ analysis: nonlinear systems with a number of observable variables ranging from 5 to 12 and a random  number of hidden confounders (from 1 to 3). (a) False Positive Rate (FPR); (b) Structural Hamming Distance (SHD); (c) Uncertainty; (d) $F_1$-Score; (e) PAG Size (reported in logarithmic scale); (f) Time (expressed in seconds).}
    \label{fig:exp_S4}
\end{figure}

\begin{figure}
    \centering
    \begin{subfigure}{.475\textwidth}
        \includegraphics[trim={0cm 0cm 1.2cm 0cm}, clip, width=\textwidth]{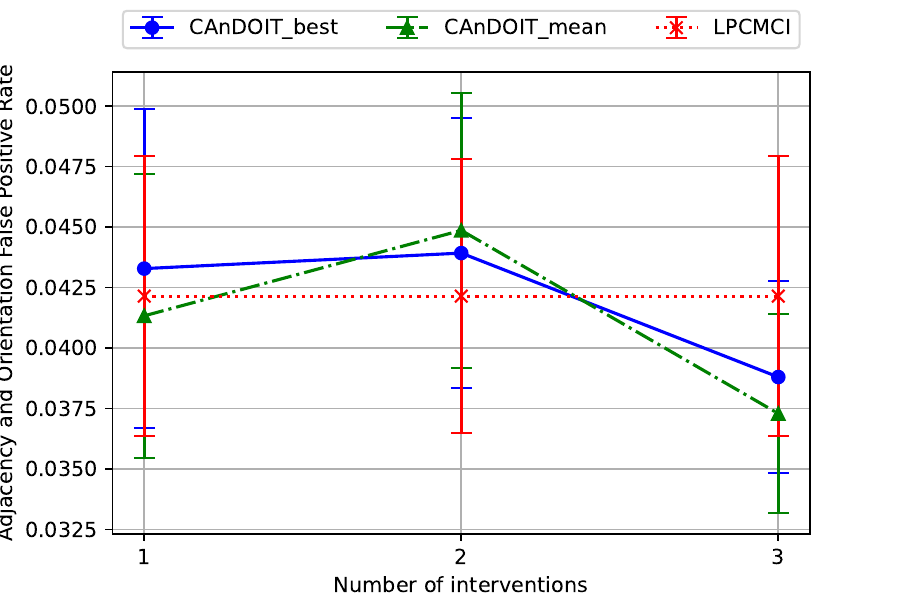}\caption{}\label{fig:s5_fpr}
    \end{subfigure}
    \qquad
    \begin{subfigure}{.475\textwidth}
        \includegraphics[trim={0.35cm 0cm 1.2cm 0cm}, clip, width=\textwidth]{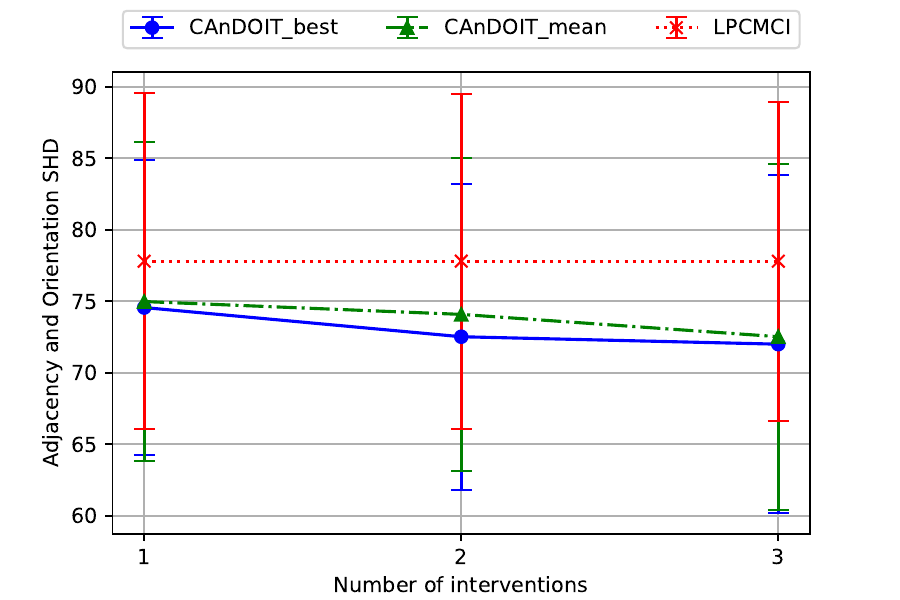}\caption{}\label{fig:s5_shd}
    \end{subfigure}\\
    \begin{subfigure}{.475\textwidth}
        \includegraphics[trim={0.35cm 0cm 1.2cm 1.2cm}, clip, width=\textwidth]{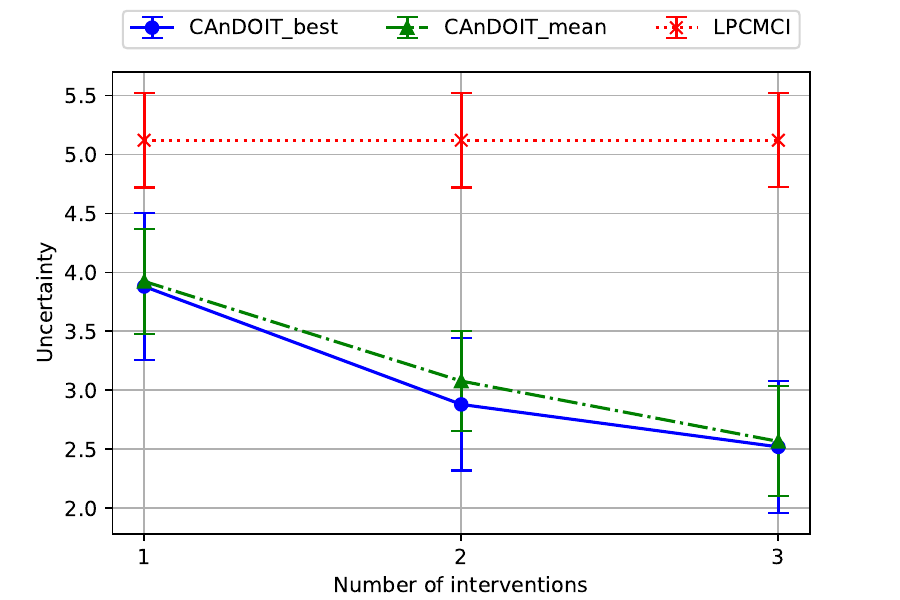}\caption{}\label{fig:s5_uncertainty}
    \end{subfigure}
    \qquad
    \begin{subfigure}{.475\textwidth}
        \includegraphics[trim={0.1cm 0cm 1.2cm 1.1cm}, clip, width=\textwidth]{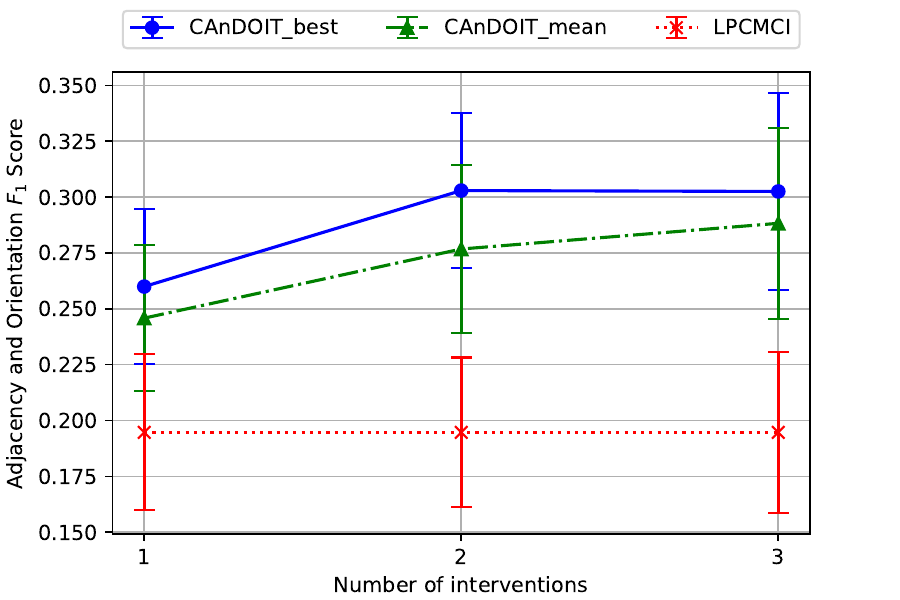}\caption{}\label{fig:s5_f1}
    \end{subfigure}\\
    \begin{subfigure}{.475\textwidth}
        \includegraphics[trim={0.35cm 0cm 1.2cm 1.2cm}, clip, width=\textwidth]{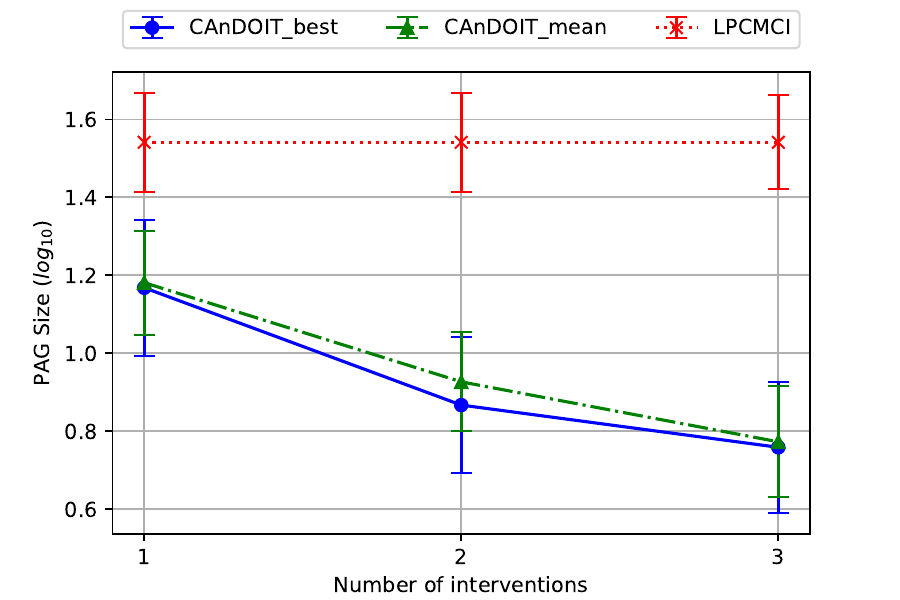}\caption{}\label{fig:s5_pag}
    \end{subfigure}
    \qquad
    \begin{subfigure}{.475\textwidth}
        \includegraphics[trim={0.35cm 0cm 1.2cm 1.1cm}, clip, width=\textwidth]{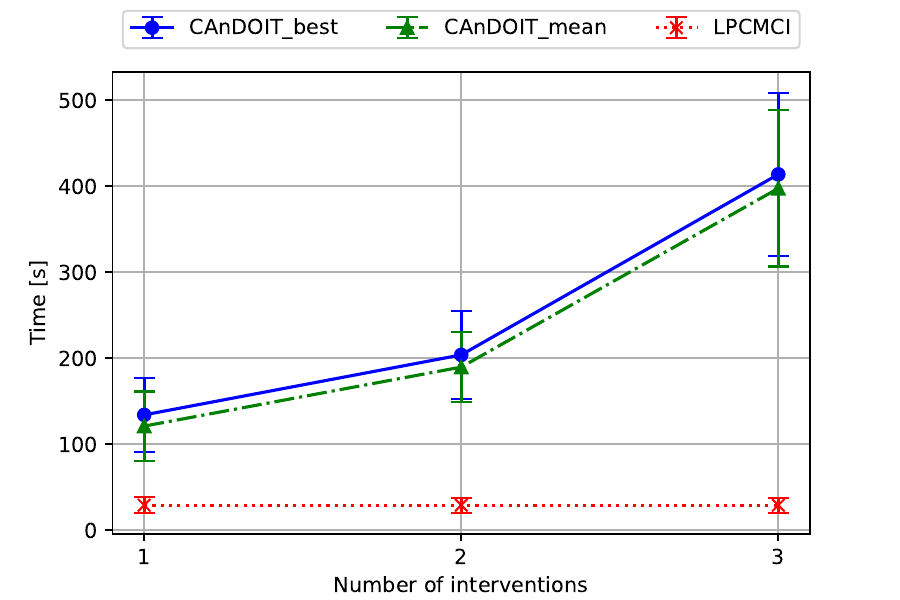}\caption{}\label{fig:s5_time}
    \end{subfigure}\\
    \caption{LPCMCI (red dotted line), CAnDOIT\_mean (green dashed line) and CAnDOIT\_best (blue) in $S_5$ analysis: nonlinear systems with 5 observable variables, a random  number of hidden confounders (from 1 to 3) and an increasing number of interventions, ranging from 1 to 3. (a) False Positive Rate (FPR); (b) Structural Hamming Distance (SHD); (c) Uncertainty; (d) $F_1$-Score; (e) PAG Size (reported in logarithmic scale); (f) Time (expressed in seconds).}
    \label{fig:exp_S5}
\end{figure}

\textbf{Figures~\ref{fig:exp_S1},~\ref{fig:exp_S2},~\ref{fig:exp_S3},~\ref{fig:exp_S4}}~and~\textbf{\ref{fig:exp_S5}} compare the causal discovery results obtained with LPCMCI~(red dotted lines), CAnDOIT\_mean~(green dashed lines), and CAnDOIT\_best~(blue lines).
The different markers in the graphs represent the mean scores computed across 25 run tests for each configuration, while the error bars show the confidence levels determined by 1000 bootstraps over the 25 results. Additionally, a Linear Mixed Model~(LMM) was used to assess the statistical validity and robustness of the analysis,\mycite{jiang2007linear} detailed results of which are provided by \textbf{Tables~\ref{table:LMM_S123}}~and~\textbf{\ref{table:LMM_S45}} in Appendix~\ref{appx:LMM_analysis}.

The $S_1,~S_2,~S_3,~S_4$ and $S_5$ analyses are presented in Figures~\ref{fig:exp_S1},~\ref{fig:exp_S2},~\ref{fig:exp_S3},~\ref{fig:exp_S4}, and~\ref{fig:exp_S5}, respectively, demonstrating the superior performance of CAnDOIT compared to LPCMCI across all scenarios. Note that, for all the analyses, the PAG Size score is shown on a logarithmic scale to improve readability.
In $S_1$ (Figure~\ref{fig:exp_S1}), CAnDOIT (both CAnDOIT\_mean and CAnDOIT\_best) performs remarkably well in all metrics. \textbf{Figures~\ref{fig:s1_fpr}},~\textbf{\ref{fig:s1_uncertainty}}~and~\textbf{\ref{fig:s1_pag}} show FPR, Uncertainty and PAG Size, respectively. It consistently maintains lower FPR and Uncertainty scores compared to LPCMCI, leading to improved identifiability across all $S_1$ cases, largely due to its use of interventional data. The superiority of CAnDOIT is evident also in the SHD and $F_1$-Score comparisons shown in \textbf{Figures~\ref{fig:s1_shd}}~and~\textbf{\ref{fig:s1_f1}}, respectively, where CAnDOIT continues to consistently outperform LPCMCI.
In terms of execution time, depicted in \textbf{Figure~\ref{fig:s1_time}}, CAnDOIT appears slower compared to LPCMCI due to the introduction of the context node used to model the single intervention.
The decreasing trend in the FPR score observed in \textbf{Figure~\ref{fig:s1_fpr}} is associated to the increasing number of variables. This decrease is a result of the higher True Negative score (TN), which grows with the number of variables.
Furthermore, it is important to specify that the different performances between CAnDOIT\_mean and CAnDOIT\_best are due to the fact that CAnDOIT\_best (the CAnDOIT result with the single intervention yielding the highest $F_1$-Score) is often associated with an intervention on a variable that is the origin of multiple ambiguous links. Intervening on such a variable is more likely to yield better performance compared to intervening on a variable that originates only a single ambiguous link. On the other hand, CAnDOIT\_mean considers all conducted interventions and averages the results.

In the $S_2$ analysis (Figure~\ref{fig:exp_S2}), which is analogous to $S_1$ but includes a random number of hidden confounders, CAnDOIT continues to outperform LPCMCI. As shown in \textbf{Figures~\ref{fig:s2_fpr}}~and~\textbf{\ref{fig:s2_uncertainty}}, CAnDOIT consistently maintains low FPR and Uncertainty scores, surpassing LPCMCI. The $S_2$ analysis reaffirms the benefits of using interventional data to enhance the identifiability of the causal graph. In particular, the PAG Size metric in \textbf{Figure~\ref{fig:s2_pag}} shows how CAnDOIT, by utilizing interventional data, is able to keep the number of equivalent MAGs consistently lower than LPCMCI. Unlike in $S_1$, the difference between CAnDOIT and LPCMCI is less pronounced in the SHD score, as shown in \textbf{Figure~\ref{fig:s2_shd}}, though the advantage of using interventional data remains evident in the $F_1$-Score (\textbf{Figure~\ref{fig:s2_f1}}). Similar to $S_1$, the employment of context variables to model interventions makes CAnDOIT slower compared to LPCMCI, as shown in \textbf{Figure~\ref{fig:s2_time}}.

In the $S_3$ analysis (Figure~\ref{fig:exp_S3}), we fixed the number of observable variables and varied the number of interventions to evaluate how our algorithm handles multiple interventions and whether they contribute to better performance compared to the single intervention cases analyzed in $S_1$ and $S_2$. CAnDOIT continues to significantly outperform LPCMCI. As shown in \textbf{Figures~\ref{fig:s3_fpr}},~\textbf{\ref{fig:s3_uncertainty}~and~\textbf{\ref{fig:s3_pag}}}, CAnDOIT yields lower FPR, Uncertainty, and PAG Size scores than LPCMCI, demonstrating how an increasing number of interventions helps reduce the number of uncertain links and, consequently, the PAG size. \textbf{Figures~\ref{fig:s3_shd}}~and~\textbf{\ref{fig:s3_f1}} show an improvement in the SHD and $F_1$ scores when transitioning from one to two interventions, though there is a slight deterioration in performance with three interventions. This can be attributed to the number of samples associated with each intervention. As explained in \textbf{Section~\ref{subsec:toy_evaluation}}, to ensure a fair comparison, LPCMCI and CAnDOIT always use the same amount of data. In the case of multiple interventions, the 300 interventional samples are equally divided among the number of interventions, meaning that with three interventions, each has 100 samples. This reduced sample size for each intervention is not sufficient to further improve the SHD and $F_1$ scores compared to the two interventions case, explaining the deterioration in these metrics in the three interventions case. As in previous analyses, CAnDOIT appears slower compared to LPCMCI, as shown in \textbf{Figure~\ref{fig:s3_time}}.

The $S_4$ analysis (Figure~\ref{fig:exp_S4}) is the nonlinear counterpart to $S_2$. Specifically, it analyzes nonlinear systems with a number of observable variables ranging from 5 to 12 and a random number of hidden confounders (from 1 to 3). Like $S_2$, this analysis helps to understand how CAnDOIT scales with an increasing number of variables in the system, but in this case with nonlinear relationships. In this analysis, CAnDOIT's FPR and SHD scores, depicted in  \textbf{Figures~\ref{fig:s4_fpr}}~and~\textbf{\ref{fig:s4_shd}} respectively, remain comparable to LPCMCI. However, the benefit of using interventional data is evident in the Uncertainty and consequently the PAG Size, as shown in \textbf{Figures~\ref{fig:s4_uncertainty}}~and~\textbf{\ref{fig:s4_pag}} respectively. Additionally, in terms of $F_1$-Score (\textbf{Figure~\ref{fig:s4_f1}}), CAnDOIT consistently outperforms LPCMCI, with both CAnDOIT\_mean and CAnDOIT\_best. The difference in terms of execution time between LPCMCI and CAnDOIT (\textbf{Figure~\ref{fig:s4_time}}) is more pronounced for the nonlinear case. This is due to conditional independence test used in this case, namely Gaussian Process and Distance Correlation (GPDC), which is slower compared to the partial correlation test used in the linear case. In CAnDOIT, the introduction of context variables to model interventions further accentuates this aspect.

Finally, in the $S_5$ analysis (Figure~\ref{fig:exp_S5}), similar to $S_3$, we fixed the number of observable variables and varied the number of interventions to assess CAnDOIT's ability to handle multiple interventions and their contribution to performance enhancement compared to a single intervention. In this case, the FPR scores for LPCMCI and CAnDOIT are comparable across the three scenarios, as shown in \textbf{Figure~\ref{fig:s5_fpr}}. However, by using interventional data, CAnDOIT outperforms LPCMCI in the remaining metrics, demonstrating its capability to remove ambiguities: it achieves lower Uncertainty and consequently lower PAG Size scores (\textbf{Figures~\ref{fig:s5_uncertainty}}~and~\textbf{\ref{fig:s5_pag}}). Performing multiple interventions results in improvements in CAnDOIT's SHD and $F_1$ scores (\textbf{Figures~\ref{fig:s5_shd}}~and~\textbf{\ref{fig:s5_f1}}) when transitioning from one to two interventions. However, as seen in $S_3$, CAnDOIT is not able to further enhance its performance with three interventions. In this case, CAnDOIT's execution time is longer compared to LPCMCI, and \textbf{Figure~\ref{fig:s5_time}} shows that this increase in time directly depends on the number of interventions.

It is important to highlight the significance of these analyses. Although in all these evaluation strategies, CAnDOIT utilized a subset of observational data alongside interventional data specifically aimed at testing links affected by ambiguities, this still yielded a consistent improvement in overall accuracy of the causal analysis. This improvement is evident in Figures~\ref{fig:s1_f1},~\ref{fig:s2_f1},~\ref{fig:s3_f1},~\ref{fig:s4_f1},~and~\ref{fig:s5_f1}, which represent the $F_1$-Scores for $S_1,~S_2,~S_3,~S_4$ and $S_5$, respectively. Additionally, there is a significant enhancement in the identifiability of the causal graph, as shown in Figures~\ref{fig:s1_pag},~\ref{fig:s2_pag},~\ref{fig:s3_pag},~\ref{fig:s4_pag},~and~\ref{fig:s5_pag}, which illustrate the PAG Size. These positive results are even more noteworthy considering that CAnDOIT operates with 300 fewer observational data samples compared to LPCMCI.

Overall, these evaluation strategies demonstrate CAnDOIT's effectiveness in handling interventional data and its superiority over the current state-of-the-art LPCMCI causal discovery algorithm in producing more accurate causal models, decreasing the uncertainty and the PAG size. The LMM analysis confirmed the statistical significance of our evaluation strategies, as indicated by the F-statistics and p-values for all scores shown in Tables~\ref{table:LMM_S123}~and~\ref{table:LMM_S45} of Appendix~\ref{appx:LMM_analysis}. Moreover, the table provides further useful statistical information, confirming the importance of the algorithm choice (LPCMCI or CAnDOIT).

\section{Evaluation on Robotic Scenario}\label{sec:causalworld_eval}
Once established, through the evaluation strategies presented in Section~\ref{sec:randomdag}, that our approach works correctly, we used it for modeling a robotic scenario in a simulated environment. Our strategy was first to extract time-series data from the simulator, and then use it for causal discovery in the presence of a hidden confounder.

\subsection{Causal World for Robot Camera Modeling}\label{subsec:causalworld}
\begin{figure}
    \centering
    \begin{subfigure}{.475\textwidth}
        \includegraphics[trim={2cm 2cm 2cm 2cm}, clip, width=\textwidth]{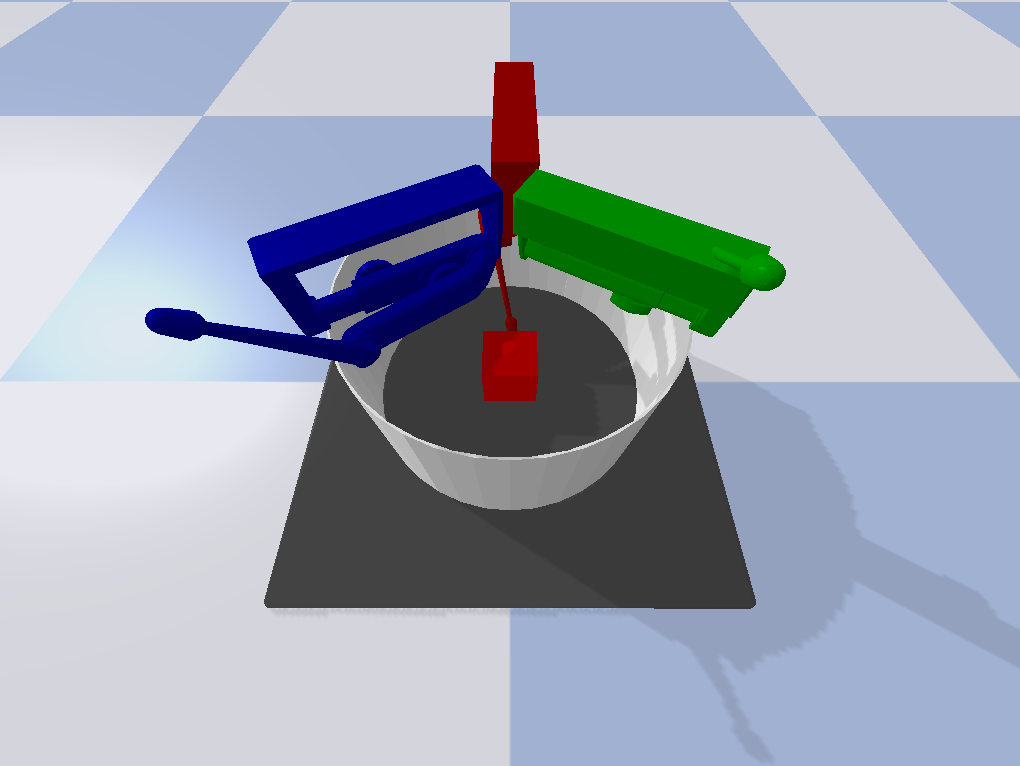}\caption{}\label{fig:CW_observation}
    \end{subfigure}
    \qquad
    \begin{subfigure}{.475\textwidth}
        \includegraphics[trim={2cm 2cm 2cm 2cm}, clip, width=\textwidth]{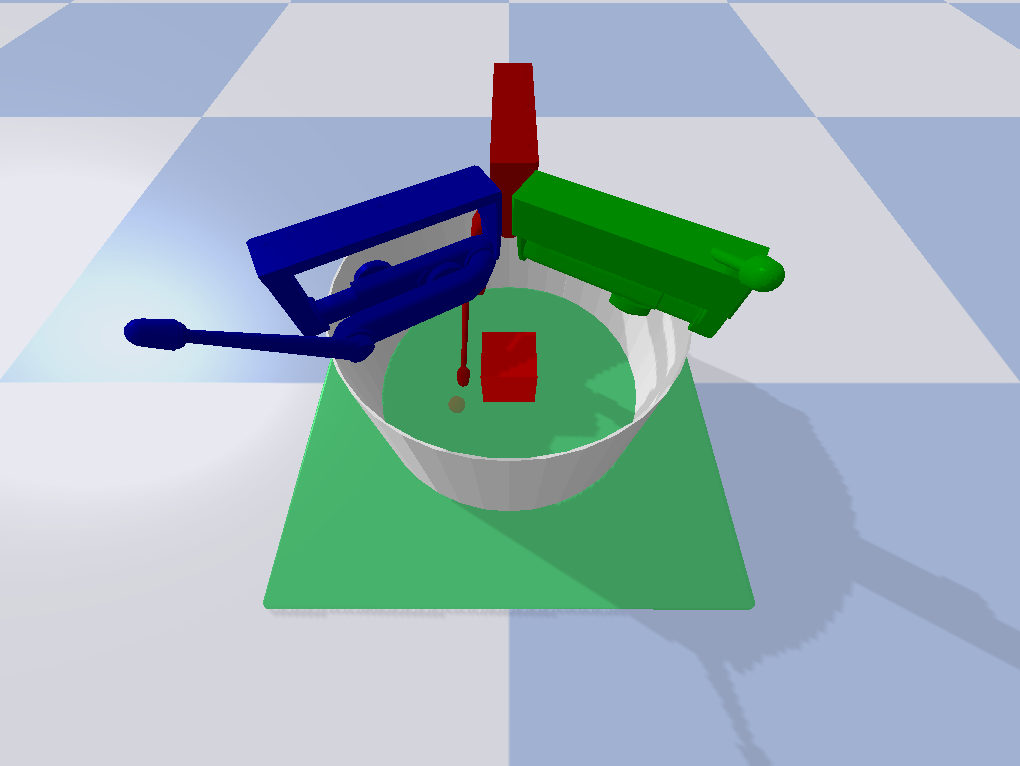}\caption{}\label{fig:CW_intervention}
    \end{subfigure}
    \caption{Causal World:\mycite{ahmed2020causalworld} a robotic manipulation environment. (left) observational experiment; (right) experiment with intervention performed on the floor's color.}\label{fig:example_causalworld}
\end{figure}
We designed an experiment to learn the causal model in a hypothetical robot arm application equipped with a camera. Our focus was on estimating the causal relationship between the color's brightness of objects as captured by the camera and various factors, including camera-to-object distance.
For this evaluation, we utilised the well-known benchmark {\em Causal World},\mycite{ahmed2020causalworld} which is designed for causal structure learning in a robotic manipulation environment. The environment consists of a TriFinger robot, a floor, and a stage. It allows for the inclusion of objects with various shapes, e.g. cubes. This simulator is widely used in the causality community due to its ability to support diverse manipulation tasks and interventions,\mycite{lachapelle2022disentanglement,yao2021learning,lippe2023biscuit} including changing the objects' color or mass.

For simplicity, we focused on a specific scenario using only one finger of the robot, where the finger's end-effector was equipped with a camera. The scenario (shown in \textbf{Figure~\ref{fig:example_causalworld}}) consists of a cube placed at the center of the floor, surrounded by a white stage. 
The color's brightness ($b$) of the cube and the floor is modeled as follows:
\begin{equation}
    b = K_h\frac{H}{H_{max}} + K_v \left(1 - \frac{v}{v_{max}}\right) + K_d\frac{d_c}{d_{c_{max}}}
\end{equation}
where $H$ is the end-effector height, $v$ its absolute velocity, and $d_c$ the distance between
the end-effector and the cube. $K_h$, $K_v$, $K_d$ are the gains associated to each factors, while $H_{max}$, $v_{max}$, and $d_{c_{max}}$ are the maximum values for $H$, $v$, and $d_c$, respectively. This model captures the shading and blurring effects on the cube due to the height of the end-effector, its velocity, and its distance from the cube. On the other hand, the floor, being darker and larger than the cube, is only affected by the end-effector's height.

The data collected from the scenario therefore includes the floor~($F_c$) and the cube~($C_c$) colors, as well as the height~($H$), the absolute velocity~($v$) of the end-effector, and its distance from the cube~($d_c$). The ground-truth structural causal model for the variables $F_c$ and $C_c$ is expressed as follows:

\begin{equation}
\begin{cases}
F_c(t) = b(H(t-1))\\
C_c(t) = b(H(t-1), v(t-1), d_c(t-1))
\end{cases}
\label{eq:causal_world_spuious_link}
\end{equation}

Note that $H$, $v$, and $d_c$ are obtained directly from the simulator and not explicitly modeled. 

\subsection{Experimental Results on Robotic Scenario}

\begin{figure}[t]
    \centering
    \begin{subfigure}{.25\textwidth}
        \includegraphics[trim={6cm 2.05cm 7cm 0.7cm}, clip, width=\textwidth]{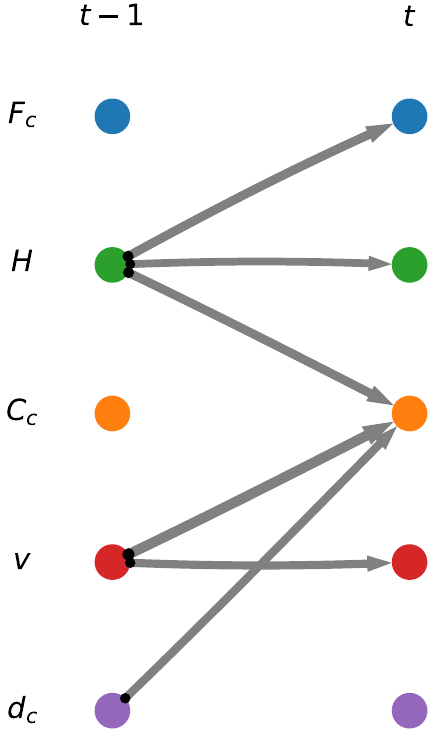}
        \caption{}\label{fig:lpcmci_complete}
    \end{subfigure}
    \begin{subfigure}{.365\textwidth}
        \includegraphics[trim={3.45cm 2cm 6cm 0.3cm}, clip, width=\textwidth]{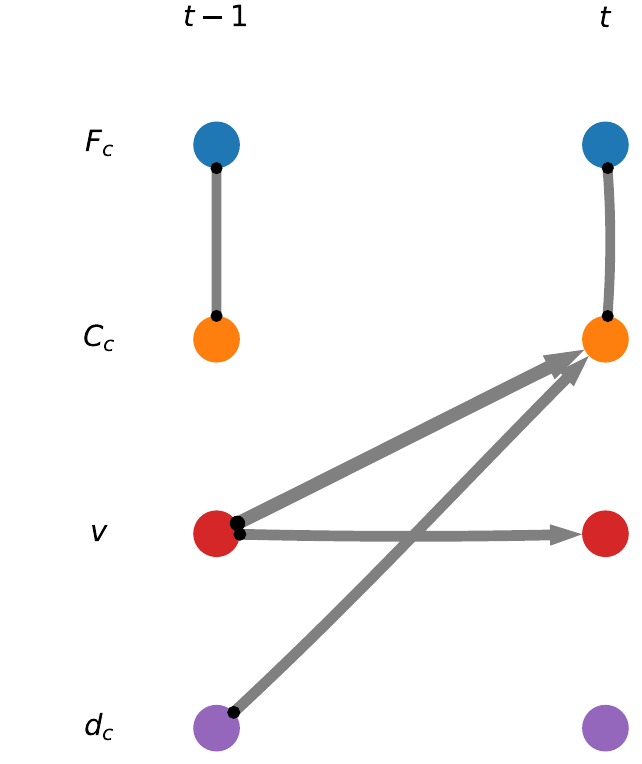}
        \caption{}\label{fig:lpcmci_hidden}
    \end{subfigure}
    \begin{subfigure}{.365\textwidth}
        \includegraphics[trim={3.45cm 2cm 6cm 0.3cm}, clip, width=\textwidth]{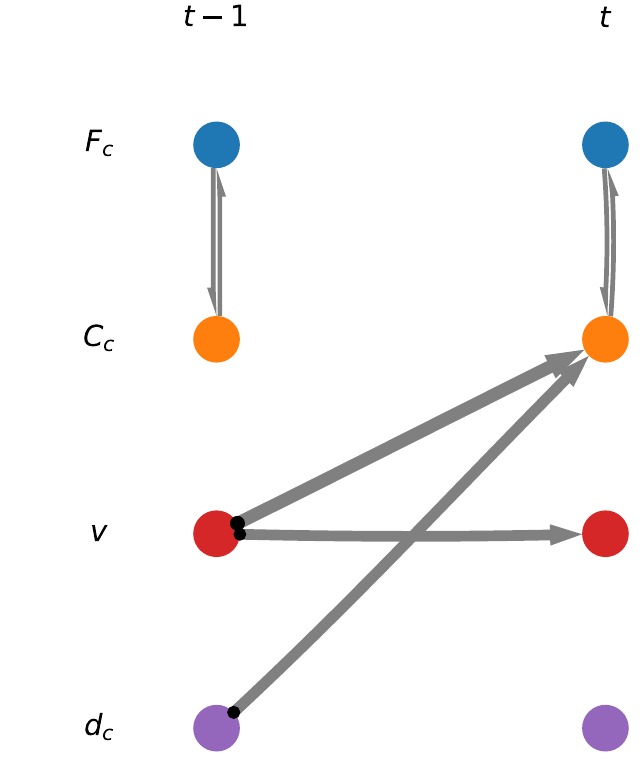}
        \caption{}\label{fig:candoit}
    \end{subfigure}
    \caption{Causal model of the robot camera in Causal World: (a) LPCMCI's result with all the variables being observable; (b) LPCMCI's result with hidden $H$; (c) CAnDOIT's causal model with hidden $H$.}
    \label{fig:exp_causal_world}
\end{figure}
The evaluation involved three main steps. {\em (i)} We generated observational data containing all the variables in the system ($F_c$, $C_c$, $H$, $v$, $d_c$), as shown in \textbf{Figure~\ref{fig:CW_observation}}, and performed the causal analysis using LPCMCI.
{\em (ii)} We intentionally hid the variable $H$, representing the height of the end-effector, to create a hidden confounder and a spurious relationship between $C_c$ and $F_c$. Again, we used LPCMCI for the causal analysis. {\em (iii)} We conducted an intervention on the floor's color, setting it to green (\textbf{Figure~\ref{fig:CW_intervention}}), and collected data from the simulator. Then we used CAnDOIT for the causal analysis with both observational and interventional, accounting for the hidden confounder $H$.
The observational time-series had a length of $600$ samples, while the interventional time-series consisted of $125$ samples. Both were recorded at a sampling rate of $10$~Hz. Also in this case, to ensure a fair analysis, LPCMCI and CAnDOIT used exactly the same amount of data. 
Consequently, LPCMCI received the complete set of observational data, whereas for CAnDOIT part of the observational data was replaced by interventions, specifically $475$ observational samples and $125$ interventional ones.

\textbf{Figure~\ref{fig:exp_causal_world}} shows the results for each specific step: (\textbf{\ref{fig:lpcmci_complete}}) causal model using LPCMCI with observable variables only; (\textbf{\ref{fig:lpcmci_hidden}}) LPCMCI's result with hidden $H$. \textbf{Figure~\ref{fig:candoit}} shows the causal model retrieved by CAnDOIT that, by leveraging both observational and interventional data, successfully identify the bidirected relation between $C_c$ and $F_c$, representing the presence of a latent confounder ($H$).

Also in this experiment, we can see the benefit of using intervention data alongside the observations. In Figure~\ref{fig:lpcmci_hidden}, LPCMCI is not able to orient the contemporaneous (spurious) link between $F_c$ and $C_c$ due to the hidden confounder $H$. This yields the ambiguous link $F_c \circlecircle C_c$, which does not encode the correct link $\leftrightarrow$ (the $\circlecircle$ represents either $\rightarrow$ or $\leftarrow$). Instead CAnDOIT, using interventional data, correctly identifies the bidirected link $F_c \leftrightarrow C_c$, decreasing once again the uncertainty level and increasing the accuracy of the reconstructed causal model.

\section{Conclusions and Future Work}\label{sec:conclusion}
In this paper, we proposed CAnDOIT, a new state-of-the-art algorithm that enables causal discovery using both observational and interventional data via context variables. We validated our approach experimentally on random synthetic models and tested on a robotic simulator for causal discovery, focusing on the significance of interventional data. Our results confirmed that CAnDOIT significantly improves previous causal discovery methods, offering enhanced accuracy, model identifiability. They also highlight its capability to handle interventional data effectively and its potential for real-world robot applications.
The proposed method lays the foundation for new observations- and interventions-based causal discovery methods on time-series data, with numerous opportunities for future research.
Firstly, to the best of our knowledge, various causal discovery algorithms combine observational data and hard/soft interventions with known/unknown targets, but only in the static domain. We present, for the first time, a causal discovery algorithm that combines observational and interventional data in a time-series setting. We acknowledge that the current version of CAnDOIT only deals with hard interventions on known targets. However, this challenge has not yet been investigated in the time-series domain. Furthermore, this solution is useful in many applications where intervention variables are known (e.g., robotics). We believe that CAnDOIT is a significant step in the right direction and future extensions could accommodate soft interventions and unknown targets.

An interesting analysis would be to test how the ratio between the lengths of observational and interventional time-series data influences the algorithm's performance. Since our application targets are intelligent robot applications, particularly in industrial and intralogistics settings,\mycite{Ghidoni2022} it would be helpful to determining the optimal amount of observational and interventional data required to establish an accurate causal model, within a reasonable time.

Scalability is another aspect to be further investigated. In this paper, we partially explored the scalability of CAnDOIT through the $S_1, S_2, S_4$ analyses, studying how the algorithm behaves with an increasing number of variables. In the future we would like to extend this analysis to check also how CAnDOIT scales with varying lengths of observational and interventional time-series data.

Finally, inspired by recent advances in the literature,\mycite{aglietti2023constrained} we plan to enhance CAnDOIT by automating the selection of intervention variables and optimising the causal model under constraints.
Also, since the current version of the algorithm first processes observational data, and then interventional one, a future improvement would be to inject opportune interventions \emph{between} observation sequences. Such a functionality would help CAnDOIT better adapt to real-world applications, where a robot observes the scenario, intervenes, and then observes again.

\medskip
\textbf{Acknowledgements} \par 
This work has received funding from the European Union's Horizon 2020 research and innovation programme under grant agreement No 101017274 (DARKO).

\bibliographystyle{unsrt}  
\bibliography{references}  

\begin{appendices}
\renewcommand{\thetable}{\Alph{section}\arabic{table}} 

\section{Random-model Generator Functionalities and Setting}\label{appx:RandomDAG_setting}
To perform the evaluation strategies explained in Section~\ref{sec:randomdag}, we developed a random-model generator capable of creating random systems of equations with hidden confounders. This tool offers various adjustable parameters, listed as follows:
\begin{itemize}
    \setlength{\itemsep}{1pt}
    \setlength{\parskip}{0pt}
    \setlength{\parsep}{0pt}
    \item time-series length;
    \item number of observable variables;
    \item number of observable parents per variable (link density);
    \item number of hidden confounders;
    \item number of confounded variables per hidden confounder;
    \item noise configuration, e.g. Gaussian noise $\mathcal{N}(\mu, \sigma^2)$;
    \item minimum $\tau_{min}$ and maximum $\tau_{max}$ time delay to consider in the equations;
    \item coefficient range of the equations' terms;
    \item functional forms applied to the equations' terms: $[-, \sin, \cos, \text{abs}, \text{pow}, \text{exp}]$, where $-$ stands for none;
    \item operators used to link various equations terms: $[+, -, *, /]$.
\end{itemize} 

For each observable variable, an equation is built by deciding a random number of parents in the set [0, link density]. Based on this number, the parents are randomly chosen among the observable variable, and all the terms are firstly built by randomly choosing a functional form and a coefficient for each of them, and then linked together by randomly choosing one of the possible operator (including noise). 
After that, a number of hidden confounders is set based on the corresponding parameter, and for each of them, a number of confounded variables is set based on the parameter. If the latter is not set, then a random number of confounded variables is picked among [0, number of observable variables parameter]. Once chosen, the hidden confounder term, comprising the functional form and the coefficient, is included in the confounded variable equation by using again one of the available operator.
Note that the influence and strength of the confounder are comparable to all other relationships within the generated system. This equality is ensured as the coefficient and functional dependency are randomly chosen from the same distribution. Specifically, the confounder connects to the confounded variables through an additive/multiplicative relationship with a random coefficient and a random functional form. \textbf{Table~\ref{tab:linear_experimental_settings}} and \textbf{Table~\ref{tab:nonlinear_experimental_settings}} display the random-model generator parameters used for the $S_1, S_2, S_3$ and $S_4, S_5$ analyses, respectively.

The parameters choice was inspired by the PCMCI, PCMCI\textsuperscript{+} and LPCMCI evaluation analyses in the literature.\mycite{runge_causal_2018,runge2020discovering,gerhardus2020high} Indeed, we used a similar noise configuration and the same coefficient range, i.e., $[0.1, 0.5]$. Moreover, differently from PCMCI\textsuperscript{+} and LPCMCI's evaluation strategy,\mycite{runge2020discovering,gerhardus2020high} where the functional forms were either linear or a predefined non-linear function, i.e. $f(x) = (1 + 5xe^{-x^2/20})x$, we employed several functional forms, both linear and non-linear, randomly built. This approach increases the sparsity and the diversification of the experiments, exploring a broader range of possible scenarios.
\renewcommand{\arraystretch}{1.1} 
\begin{table}[t]
\centering
\caption{$S_1, S_2$ and $S_3$ experimental settings}
\label{tab:linear_experimental_settings}
\begin{tabular}{|l|c|c|c|}
\hline
& \textbf{$S_1$ evaluation strategy} & \textbf{$S_2$ evaluation strategy} & \textbf{$S_3$ evaluation strategy} \\ \hline
Observational ts length & \makecell{$1300$ [LPCMCI]\\$1000$ [CAnDOIT]} & \makecell{$1300$ [LPCMCI]\\$1000$ [CAnDOIT]} & \makecell{$1300$ [LPCMCI]\\$1000$ [CAnDOIT]} \\
Interventional ts length & 300 [CAnDOIT] & 300 [CAnDOIT] & 300 [CAnDOIT] \\
N. of observable variables & from 5 to 12 & from 5 to 12 & 5 \\
N. of interventions & 1 & 1 & from 1 to 3 \\
Link density & 3 & 3 & 3 \\
Hidden confounders & 0 & rand(1, 3) & rand(1, 3) \\
Confounded variables & 0 & 3 & 3 \\
Noise configuration & \makecell{random choice among \\ $x = \text{rand}(-0.5, 2)$\\\(\text{Uniform}(-x, x)\) \\ \(\mathcal{N}(0, x)\) \\ \(\text{Weibull}(2, 1)\)\mycite{gerhardus2020high}} & \makecell{random choice among \\ $x = \text{rand}(-0.5, 2)$\\\(\text{Uniform}(-x, x)\) \\ \(\mathcal{N}(0, x)\) \\ \(\text{Weibull}(2, 1)\)\mycite{gerhardus2020high}} & \makecell{random choice among \\ $x = \text{rand}(-0.5, 2)$\\\(\text{Uniform}(-x, x)\) \\ \(\mathcal{N}(0, x)\) \\ \(\text{Weibull}(2, 1)\)\mycite{gerhardus2020high}} \\
Min time-step lag & $0$ & $0$ & $0$ \\
Max time-step lag & $3$ & $3$ & $3$ \\
Coefficient range & $[0.1, 0.5]$ & $[0.1, 0.5]$ & $[0.1, 0.5]$ \\
Operators & $[+, -]$ & $[+, -]$ & $[+, -]$ \\
Functional forms & - & - & - \\ \hline
\end{tabular}
\end{table}

\begin{table}[t]
\centering
\caption{$S_4$ and $S_5$ experimental settings}
\label{tab:nonlinear_experimental_settings}
\begin{tabular}{|l|c|c|}
\hline
& \textbf{$S_4$ evaluation strategy} & \textbf{$S_5$ evaluation strategy} \\ \hline
Observational ts length & \makecell{$1300$ [LPCMCI]\\$1000$ [CAnDOIT]} & \makecell{$1300$ [LPCMCI]\\$1000$ [CAnDOIT]} \\
Interventional ts length & 300 [CAnDOIT] & 300 [CAnDOIT] \\
N. of observable variables & from 5 to 12 & 5 \\
N. of interventions & 1 & from 1 to 3 \\
Link density & 3 & 3 \\
Hidden confounders & rand(1, 3) & rand(1, 3) \\
Confounded variables & 3 & 3 \\
Noise configuration & \makecell{random choice among \\ $x = \text{rand}(-0.5, 2)$\\\(\text{Uniform}(-x, x)\) \\ \(\mathcal{N}(0, x)\) \\ \(\text{Weibull}(2, 1)\)\mycite{gerhardus2020high}} & \makecell{random choice among \\ $x = \text{rand}(-0.5, 2)$\\\(\text{Uniform}(-x, x)\) \\ \(\mathcal{N}(0, x)\) \\ \(\text{Weibull}(2, 1)\)\mycite{gerhardus2020high}} \\
Min time-step lag & $0$ & $0$ \\
Max time-step lag & $3$ & $3$ \\
Coefficient range & $[0.1, 0.5]$ & $[0.1, 0.5]$ \\
Operators & $[+, -, *, /]$ & $[+, -, *, /]$ \\
Functional forms &$[-, \sin, \cos, \text{abs}, \text{pow}, \text{exp}]$ & $[-, \sin, \cos, \text{abs}, \text{pow}, \text{exp}]$ \\ \hline
\end{tabular}
\end{table}

\section{Statistical Evaluation of Synthetic Models Analysis}\label{appx:LMM_analysis}
In the main text, we assert the utilization of a Linear Mixed Model (LMM) to comprehensively assess the validity and robustness of our evaluation strategies. Subsequently, this section provides additional insights and commentary on the statistical evaluation, accompanied by Table~\ref{table:LMM_S123} corresponding to strategies $S_1,~S_2$ and $S_3$, and Table~\ref{table:LMM_S45} corresponding to strategies $S_4$ and $S_5$. These tables encapsulate not only the LMM results but also the Estimated Marginal Means (EMM) and Eta-squared ($\eta^2$) metrics.

EMMs represent a statistical concept used to estimate the average values of a response variable at specific levels or combinations of categorical predictor variables, accounting for the effects of other covariates within the statistical model. In simpler terms, EMMs allow us to summarize and interpret the average impact of the algorithm choice (LPCMCI or CAnDOIT) on the respective score, while controlling for other variables like the number of observable variables (nvars) or the number of interventions (nint).
On the other hand, Eta-squared ($\eta^2$) quantifies the proportion of variance in the dependent variable that is elucidated by the independent variables within the statistical model. Essentially, $\eta^2$ provides crucial insights into the strength of the relationship between predictors and response variables. In our analysis, $\eta^2$ is used to quantify the proportion of variance in our scores that can be attributed to the algorithm as well as the nvars/nint factor.

Tables~\ref{table:LMM_S123} and~\ref{table:LMM_S45} present a comprehensive overview of the LMM results obtained from the $S_1,~S_2,~S_3$ and $S_4,~S_5$ analyses, respectively. The high F-statistic values, coupled with p-values consistently below the conventional threshold ($0.05$), substantiate the statistical significance and validity of the analyses conducted on the FPR, Uncertainty, PAG Size, SHD, $F_1$ scores and the execution time.
The EMMs highlighted in bold, corresponding to the best value (highest for the $F_1$-Score and lowest for all the other scores), are mainly associated with our approach, indicating superior performance, except for execution time, where LPCMCI outperformed our CAnDOIT in all analyses.

Overall, the table reinforces the conclusions drawn from Figures~\ref{fig:exp_S1},~\ref{fig:exp_S2},~\ref{fig:exp_S3},~\ref{fig:exp_S4}~and~\ref{fig:exp_S5} and underscores the assertions made in the main text: our CAnDOIT effectively handles interventional data and consistently outperforms the current state-of-the-art LPCMCI causal discovery algorithm in producing more accurate causal models, decreasing the uncertainty and the PAG size.

\renewcommand{\arraystretch}{1.1} 
\begin{table}[t]
\setlength{\tabcolsep}{4pt}
\small
\centering
\caption{LMM results for FPR, Uncertainty, PAG Size, SHD, $F_1$ scores and execution time related to the $S_1,~S_2$ and $S_3$ evaluation strategies.}
\label{table:LMM_S123}
\begin{tabular}{l|cccc|cccc|cccc}
& \multicolumn{4}{c}{\textbf{$S_1$ evaluation strategy}} & \multicolumn{4}{|c}{\textbf{$S_2$ evaluation strategy}} & \multicolumn{4}{|c}{\textbf{$S_3$ evaluation strategy}}\\ \hline
\multicolumn{1}{l|}{} & Estim. & p-value & $\eta^2$ & EMM & Estim. & p-value & $\eta^2$ & EMM & Estim. & p-value & $\eta^2$ & EMM \\ \hline
\textbf{FPR} & \multicolumn{4}{l}{\makecell[l]{F-statistic: 3.207e+04 \\ p-value: $<$ 2.2e-16}} & \multicolumn{4}{|l}{\makecell[l]{F-statistic: 5.107e+04 \\ p-value: $<$ 2.2e-16}} & \multicolumn{4}{|l}{\makecell[l]{F-statistic: 9381 \\ p-value: $<$ 2.2e-16}}\\ \hline
Intercept       & 0.029  & $<$ 2e-16 & -    & -              & 0.044  & $<$ 2e-16 & -    & -              & 3.747e-02  & $<$ 2e-16 & -        & -\\
CAnDOIT\_best   & -      & -         & 0.10 & \textbf{0.021} & -      & -         & 0.03 & \textbf{0.027} & -          & -         & 0.11     & \textbf{0.035}\\
CAnDOIT\_mean   & 0.005  & $<$ 2e-16  & 0.10 & 0.027         & 0.002  & $<$ 2e-16 & 0.03 & 0.029          & 3.923e-03  & $<$ 2e-16 & 0.11     & 0.039\\
LPCMCI          & 0.007  & $<$ 2e-16  & 0.10 & 0.029         & 0.004  & $<$ 2e-16 & 0.03 & 0.032          & 1.225e-02  & $<$ 2e-16 & 0.11     & 0.048\\
nvars/nint      & -0.001 & $<$ 2e-16  & 0.05 & -             & -0.002 & $<$ 2e-16 & 0.18 & -              & -9.445e-04 & $<$ 2e-16 & 2.79e-03 & -\\ \hline
\textbf{Uncertainty} & \multicolumn{4}{c}{\makecell[l]{F-statistic: 2.502e+05 \\ p-value: $<$ 2.2e-16}} & \multicolumn{4}{|c}{\makecell[l]{F-statistic: 2.175e+05 \\ p-value: $<$ 2.2e-16}}  & \multicolumn{4}{|c}{\makecell[l]{F-statistic: 1.139e+05 \\ p-value: $<$ 2.2e-16}}\\ \hline
Intercept       & -4.91 & $<$ 2e-16 & -    & -             & -3.99 & $<$ 2e-16 & -    & -             & 2.918  & $<$ 2e-16 & -    & -  \\
CAnDOIT\_best   & -     & -         & 0.24 & \textbf{5.77} & -     & -         & 0.21 & \textbf{6.62} & -      & -         & 0.59 & \textbf{1.86}\\
CAnDOIT\_mean   & 1.77  & $<$ 2e-16 & 0.24 & 7.54          & 1.298 & $<$ 2e-16 & 0.21 & 7.91          & 0.661  & $<$ 2e-16 & 0.59 & 2.52 \\
LPCMCI          & 4.11  & $<$ 2e-16 & 0.24 & 9.87          & 3.924 & $<$ 2e-16 & 0.21 & 10.53         & 4.231  & $<$ 2e-16 & 0.59 & 6.091 \\
nvars/nint      & 1.25  & $<$ 2e-16 & 0.48 & -             & 1.247 & $<$ 2e-16 & 0.45 & -             & -0.528 & $<$ 2e-16 & 0.07 & - \\ \hline
\textbf{PAG Size} & \multicolumn{4}{c}{\makecell[l]{F-statistic: 2.486e+05 \\ p-value: $<$ 2.2e-16}} & \multicolumn{4}{|c}{\makecell[l]{F-statistic: 2.155e+05 \\ p-value: $<$ 2.2e-16}} & \multicolumn{4}{|c}{\makecell[l]{F-statistic: 1.128e+05 \\ p-value: $<$ 2.2e-16}} \\ \hline
Intercept       & -1.478 & $<$ 2e-16 & -    & -            & -1.189 & $<$ 2e-16 & -    & -             & 0.883 & $<$ 2e-16 & -    & -\\
CAnDOIT\_best   & -     & -         & 0.24 & \textbf{1.74} & -      & -         & 0.21 & \textbf{1.99} & -      & -         & 0.59 & \textbf{0.562}\\
CAnDOIT\_mean   & 0.534 & $<$ 2e-16  & 0.24 & 2.27         & 0.383  & $<$ 2e-16 & 0.21 & 2.38          & 0.197  & $<$ 2e-16 & 0.59 & 0.76\\
LPCMCI          & 1.237 & $<$ 2e-16 & 0.24 & 2.97          & 1.166  & $<$ 2e-16 & 0.21 & 3.17          & 1.267  & $<$ 2e-16 & 0.59 & 1.83\\
nvars/nint      & 0.378 & $<$ 2e-16 & 0.48 & -             & 0.375  & $<$ 2e-16 & 0.45 & -             & -0.16  & $<$ 2e-16 & 0.07 & -\\ \hline
\textbf{SHD} & \multicolumn{4}{c}{\makecell[l]{F-statistic: 4.272e+05 \\ p-value: $<$ 2.2e-16}} & \multicolumn{4}{|c}{\makecell[l]{F-statistic: 3.079e+04 \\ p-value: $<$ 2.2e-16}} & \multicolumn{4}{|c}{\makecell[l]{F-statistic: 456 \\ p-value: $<$ 2.2e-16}}\\ \hline
Intercept      & -17.6   & $<$ 2e-16 & -    & -              & 29.993 & $<$ 2e-16 & -    & -                  & 93.575 & $<$ 2e-16 & -    & -\\
CAnDOIT\_best  & -       & -         & 0.10 & \textbf{41.46} & -      & -         & 1.44e-03 & \textbf{149.3} & -      & -         & 5.82e-03 & \textbf{92.24}\\
CAnDOIT\_mean  & 6.295   & $<$ 2e-16  & 0.10 & 47.76         & 4.504  & $<$ 2e-16 & 1.44e-03 & 153.8          & 2.981  & $<$ 2e-16 & 5.82e-03 & 95.22\\
LPCMCI         & 8.758   & $<$ 2e-16  & 0.10 & 50.22         & 7.612  & $<$ 2e-16 & 1.44e-03 & 156.9          & 6.753  & $<$ 2e-16 & 5.82e-03 & 98.99\\
nvars/nint     & 6.946   & $<$ 2e-16 & 0.67 & -              & 14.033 & $<$ 2e-16 & 0.13 & -                  & -0.67  & 6.35e-13  & 2.30e-04 & -  \\ \hline 
\textbf{$F_1$-Score} & \multicolumn{4}{c}{\makecell[l]{F-statistic: 2.183e+04 \\ p-value: $<$ 2.2e-16}} & \multicolumn{4}{|c}{\makecell[l]{F-statistic: 7871 \\ p-value: $<$ 2.2e-16}} & \multicolumn{4}{|c}{\makecell[l]{F-statistic: 3402 \\ p-value: $<$ 2.2e-16}}\\ \hline 
Intercept      & 0.626  & $<$ 2e-16 & -    & -              & 0.297   & $<$ 2e-16 & -        & -              & 0.369   & $<$ 2e-16 & -        & -\\
CAnDOIT\_best  & -      & -         & 0.08 & \textbf{0.563} & -       & -         & 0.04     & \textbf{0.315} & -       & -         & 0.04     & \textbf{0.3817}\\
CAnDOIT\_mean  & -0.056 & $<$ 2e-16 & 0.08 & 0.507          & -0.0338 & $<$ 2e-16 & 0.04     & 0.281          & -0.030 & $<$ 2e-16 & 0.04     & 0.3519\\
LPCMCI         & -0.084 & $<$ 2e-16 & 0.08 & 0.479          & -0.050  & $<$ 2e-16 & 0.04     & 0.265          & -0.055  & $<$ 2e-16 & 0.04     & 0.3272\\
nvars/nint     & -0.007 & $<$ 2e-16 & 0.02 & -              & 0.002   & $<$ 2e-16 & 2.00e-03 & -              & 0.006   & $<$ 2e-16 & 2.19e-03 & -\\ \hline
\textbf{Time} & \multicolumn{4}{c}{\makecell[l]{F-statistic: 5293 \\ p-value: $<$ 2.2e-16}} & \multicolumn{4}{|c}{\makecell[l]{F-statistic: 7810 \\ p-value: $<$ 2.2e-16}} & \multicolumn{4}{|c}{\makecell[l]{F-statistic: 2012 \\ p-value: $<$ 2.2e-16}} \\ \hline 
Intercept      & -4.727 & $<$ 2e-16 & -    & -              & -4.975 & $<$ 2e-16 & -    & -             & 15.847 & $<$ 2e-16 & -        & - \\
CAnDOIT\_best  & -      & -         & 0.01 & 7.91           & -      & -         & 0.01 & 13.05         & -      & -         & 0.02     & 22.38\\
CAnDOIT\_mean  & 3.643  & $<$ 2e-16 & 0.01 & 11.55          & -2.201 & $<$ 2e-16 & 0.01 & 10.63         & -2.445 & 1.22e-15  & 0.02     & 19.94 \\
LPCMCI         & -3.962 & $<$ 2e-16 & 0.01 & \textbf{3.94}  & -7.507 & $<$ 2e-16 & 0.01 & \textbf{5.37} & -20.828 & $<$ 2e-16 & 0.02     & \textbf{1.55}          \\
nvars/nint     & 1.486  & $<$ 2e-16 & 0.01 & -              & 2.108  & $<$ 2e-16 & 0.03 & -             & 3.271  & $<$ 2e-16 & 2.04e-03 & - \\ \hline
\end{tabular}
\end{table}

\begin{table}[t]
\setlength{\tabcolsep}{4pt}
\small
\centering
\caption{LMM results for FPR, Uncertainty, PAG Size, SHD, $F_1$ scores and execution time related to the $S_4$ and $S_5$ evaluation strategies.}
\label{table:LMM_S45}
\begin{tabular}{l|cccc|cccc}
& \multicolumn{4}{c}{\textbf{$S_4$ evaluation strategy}} & \multicolumn{4}{|c}{\textbf{$S_5$ evaluation strategy}}\\ \hline
\multicolumn{1}{l|}{} & Estim. & p-value & $\eta^2$ & EMM & Estim. & p-value & $\eta^2$ & EMM \\ \hline
\textbf{FPR} & \multicolumn{4}{l}{\makecell[l]{F-statistic: 4.107e+04 \\ p-value: $<$ 2.2e-16}} & \multicolumn{4}{|l}{\makecell[l]{F-statistic: 558 \\ p-value: $<$ 2.2e-16}} \\ \hline
Intercept      & 4.431e-02  & $<$ 2e-16 & -        & -                & 4.479e-02 & $<$ 2e-16 & -    & -              \\
CAnDOIT\_best  & -          & -         & 1.86e-03 & \textbf{0.02905} & -         & -         & 1.00e-03 & 0.0420 \\
CAnDOIT\_mean  & 3.239e-04  & $<$ 2e-16 & 1.86e-03 & 0.02937          & -8.828e-04 & $<$ 2e-16 & 1.00e-03 & \textbf{0.0411}          \\
LPCMCI         & 9.300e-04  & $<$ 2e-16 & 1.86e-03 & 0.02998          & 1.383e-04 & $<$ 2e-16 & 1.00e-03 & 0.0421          \\
nvars/nint     & -1.797e-03 & $<$ 2e-16 & 0.17     & -                & -1.396e-03 & $<$ 2e-16 & 6.40e-03 & -              \\ \hline
\textbf{Uncertainty} & \multicolumn{4}{c}{\makecell[l]{F-statistic: 2.854e+05 \\ p-value: $<$ 2.2e-16}} & \multicolumn{4}{|c}{\makecell[l]{F-statistic: 4.741e+04 \\ p-value: $<$ 2.2e-16}}  \\ \hline
Intercept      & -3.03 & $<$ 2e-16 & -    & -              & 3.99 & $<$ 2e-16 & -        & -              \\
CAnDOIT\_best  & -     & -         & 0.15 & \textbf{7.966} & -     & -         & 0.35     & \textbf{3.099} \\
CAnDOIT\_mean  & 0.217 & $<$ 2e-16 & 0.15 & 8.183          & 0.09 & $<$ 2e-16 & 0.35     & 3.189          \\
LPCMCI         & 2.472 & $<$ 2e-16 & 0.15 & 10.439         & 2.02 & $<$ 2e-16 & 0.35     & 5.123          \\
nvars/nint     & 1.293 & $<$ 2e-16 & 0.56 & -              & -0.45 & $<$ 2e-16 & 0.08 & -              \\ \hline
\textbf{PAG Size} & \multicolumn{4}{c}{\makecell[l]{F-statistic: 2.85e+05 \\ p-value: $<$ 2.2e-16}} & \multicolumn{4}{|c}{\makecell[l]{F-statistic: 4.787e+04 \\ p-value: $<$ 2.2e-16}}  \\ \hline
Intercept      & -0.909 & $<$ 2e-16 & -        & -             & 1.20 & $<$ 2.2e-16 & -    & -             \\
CAnDOIT\_best  & -     & -         & 0.15     & \textbf{2.399} & -     & -           & 0.36 & \textbf{0.92} \\
CAnDOIT\_mean  & 0.063 & $<$ 2e-16 & 0.15     & 2.463          & 0.03 & $<$ 2e-16 & 0.36 & 0.95          \\
LPCMCI         & 0.745 & $<$ 2e-16 & 0.15     & 3.144          & 0.61 & $<$ 2e-16 & 0.36 & 1.54          \\
nvars/nint     & 0.389 & $<$ 2e-16 & 0.56 & -                  & -0.13 & $<$ 2e-16 & 0.08 & -             \\ \hline
\textbf{SHD} & \multicolumn{4}{c}{\makecell[l]{F-statistic: 1.489e+04 \\ p-value: $<$ 2.2e-16}} & \multicolumn{4}{|c}{\makecell[l]{F-statistic: 409.7 \\ p-value: $<$ 2.2e-16}} \\ \hline
Intercept      & 86.828 & $<$ 2e-16 & -        & -             & 74.826 & $<$ 2e-16 & -    & -             \\
CAnDOIT\_best  & -      & -         & 7.18e-04 & \textbf{165.2} & -     & -         & 4.85e-03 & \textbf{73.08} \\
CAnDOIT\_mean  & 2.140  & $<$ 2e-16 & 7.18e-04 & 167.3         & 0.717 & 2.24e-06 & 4.85e-03 & 73.80          \\
LPCMCI         & 5.117  & $<$ 2e-16 & 7.18e-04 & 170.3         & 4.659 & $<$ 2e-16 & 4.85e-03 & 77.74         \\
nvars/nint     & 9.218  & $<$ 2e-16 & 0.07     & -             & -0.871 & $<$ 2e-16 & 5.88e-04 & -             \\ \hline 
\textbf{$F_1$-Score} & \multicolumn{4}{c}{\makecell[l]{F-statistic: 1.399e+04 \\ p-value: $<$ 2.2e-16}} & \multicolumn{4}{|c}{\makecell[l]{F-statistic: 1.455e+04 \\ p-value: $<$ 2.2e-16}} \\ \hline 
Intercept      & 0.17   & $<$ 2e-16 & -    & -              & 0.259  & $<$ 2e-16 & -    & -              \\
CAnDOIT\_best  & -      & -         & 0.04 & \textbf{0.22}  & -          & -         & 0.15 & \textbf{0.2882} \\
CAnDOIT\_mean  & -0.016 & $<$ 2e-16 & 0.04 & 0.21           & -0.017 & $<$ 2e-16 & 0.15 & 0.2705          \\
LPCMCI         & -0.039 & $<$ 2e-16 & 0.04 & 0.18           & -0.093 & $<$ 2e-16 & 0.15 & 0.1944          \\
nvars/nint     & 0.006  & $<$ 2e-16 & 0.03 & -              & 0.014 & $<$ 2e-16 & 0.01 & -              \\ \hline
\textbf{Time} & \multicolumn{4}{c}{\makecell[l]{F-statistic: 7.177e+04 \\ p-value: $<$ 2.2e-16}} & \multicolumn{4}{|c}{\makecell[l]{F-statistic: 5.28e+04 \\ p-value: $<$ 2.2e-16}} \\ \hline 
Intercept      & 38.13 & $<$ 2e-16 & -    & -               & 64.04 & $<$ 2e-16 & -    & -              \\
CAnDOIT\_best  & -        & -         & 0.15 & 267.2 & -    & -         & 0.31 & 250.35         \\
CAnDOIT\_mean  & -10.07    & $<$ 2e-16 & 0.15 & 257.2       & -12.477 & $<$ 2e-16 & 0.31 & 237.87 \\
LPCMCI         & -131.27  & $<$ 2e-16 & 0.15 & \textbf{136} & -221.22 & $<$ 2e-16 & 0.31 & \textbf{29.13}          \\
nvars/nint     & 26.96   & $<$ 2e-16 & 0.16 & -             & 93.22  & $<$ 2e-16 & 0.20 & -              \\ \hline
\end{tabular}
\end{table}
\end{appendices}

\end{document}